\documentclass[11pt]{article}
\usepackage{cite} 
\usepackage{geometry}
\usepackage{amsmath,amssymb,amsfonts} 
\usepackage{algorithmic} 
\usepackage{graphicx} 
\usepackage{algorithm,algorithmic} 
\usepackage{hyperref} 
\hypersetup{hidelinks=true} 
\usepackage{glossaries} 
\usepackage{textcomp} 
\usepackage{multirow} 
\usepackage{makecell} 
\usepackage{xurl} 
\usepackage[normalem]{ulem} 
\usepackage[table]{xcolor} 
\definecolor{good}{RGB}{0,150,0} 
\definecolor{bad}{RGB}{200,0,0}

\def\BibTeX{{\rm B\kern-.05em{\sc i\kern-.025em b}\kern-.08em
    T\kern-.1667em\lower.7ex\hbox{E}\kern-.125emX}}
\begin{document}
\title{A Comparative Study of CNN Optimization Methods for Edge AI: Exploring the Role of Early Exits}

\author{Nekane Fernandez, 
Ivan Valdes,  
Steven Van Vaerenbergh,
Idoia de la Iglesia, \\ and 
Julen Arratibel
\thanks{Nekane Fernandez, Ivan Valdes, Idoia de la Iglesia and Julen Arratibel are with the Dept. of Connected and Distributed Intelligence, Ikerlan Technology Research Centre, Arrasate, Spain 
(email: nekane.fernandez@ikerlan.es; ivaldes@ikerlan.es; idelaiglesia@ikerlan.es; jarratibel@ikerlan.es).}
\thanks{Nekane Fernandez and Steven Van Vaerenbergh are with the Dept. of Mathematics, Statistics and Computation, Universidad de Cantabria, Santander, Spain 
(email: nekane.fernandez@alumnos.unican.es; steven.vanvaerenbergh@unican.es).}}

\newacronym{ai}{AI}{Artificial Intelligence}
\newacronym{dl}{DL}{Deep Learning}
\newacronym{iot}{IoT}{Internet of Things}
\newacronym{cc}{CC}{Cloud Computing}
\newacronym{ec}{EC}{Edge Computing}
\newacronym{cnn}{CNNs}{Convolutional Neural Networks}
\newacronym{aiot}{AIoT}{Artificial Intelligence of Things}
\newacronym{dnn}{DNNs}{Deep Neural Networks}
\newacronym{kd}{KD}{Knowledge Distillation}
\newacronym{fpga}{FPGA}{Field Programmable Gate Array}
\newacronym{ptq}{PTQ}{Post-Training Quantization}
\newacronym{dq}{DQ}{Dynamic Quantization}
\newacronym{qat}{QAT}{Quantization-Aware Training}
\newacronym{threshold}{T}{threshold}
\newacronym{cg}{CG}{channel granularity}
\newacronym{pr}{PR}{pruning ratio}

\maketitle

\begin{abstract}
Deploying deep neural networks on edge devices requires balancing accuracy, latency, and resource constraints under realistic execution conditions. To fit models within these constraints, two broad strategies have emerged: static compression techniques, such as pruning and quantization, which permanently reduce model size; and dynamic approaches, such as early-exit mechanisms, which adapt computational cost at runtime. While both families are widely studied in isolation, they are rarely compared under identical conditions on physical hardware. This paper presents a unified, deployment-oriented comparison of static compression and dynamic early-exit mechanisms, evaluated on real edge devices using ONNX-based inference pipelines. Our results show that static and dynamic compression techniques offer fundamentally different trade-offs for edge deployment. While pruning and quantization deliver consistent memory footprint reduction, early-exit mechanisms enable input-adaptive computation savings that they cannot match. Their combination, however, proves transformative synergy—simultaneously reducing inference latency and memory usage with minimal accuracy loss, fundamentally expanding the boundaries of what is achievable at the edge.
\end{abstract}


\section{Introduction} \label{sec:introduction}
The integration of \gls{ai} and \gls{iot} — known as \gls{aiot} — is reshaping the deployment of intelligent services in distributed environments. By combining the pervasive sensing and connectivity of \gls{iot} with AI-driven decision-making, \gls{aiot} enables advanced applications in smart cities, industrial automation, and autonomous systems~\cite{10929047}.

The effectiveness of these applications, however, hinges on the successful deployment of \gls{dl} models. In particular, \gls{cnn} have become the benchmark for perceptual tasks central to \gls{aiot}, ranging from object detection in autonomous vehicles to defect inspection in smart manufacturing to gesture recognition in wearable devices \cite{Elhanashi2023Integration, Chen2020Deep}. While IoT devices generate vast amounts of data that often demand real-time or near-real-time \gls{ai} processing, the increasing depth and complexity of modern \gls{dl} models introduce substantial computational and memory overhead. This growing resource demand poses significant challenges for deployment in resource-constrained environments \cite{Daghero_Energy-efficient, Kartikeya_Vision, 10398463}.

A common approach to alleviating these constraints is to offload inference to the cloud. However, cloud-centric inference stands in tension with a core promise of \gls{aiot}: autonomous, low-latency decision-making at the source. Concerns over latency, privacy, reliability, and bandwidth further limit its suitability in many \gls{aiot} scenarios \cite{Satyanarayanan_Emergence}. In response, \gls{ec} has emerged as a compelling paradigm that brings computation closer to the data source — a vision often termed Edge Intelligence — reducing cloud dependency, alleviating network congestion, and enabling faster, context-aware responses \cite{Wang_Empowering, Zhou_Paving, Semerikov2025Edge}. Nevertheless, edge nodes themselves remain constrained in processing power, memory, storage, and energy capacity \cite{Luiz_Integration, electronics14173468}.

To this end, the field has developed two broad families of post-design optimization techniques. The first aims at static model compression, creating a permanently smaller or more efficient version of a model through methods like pruning and quantization, altering the model's parameters or structure before deployment. The second family focuses on dynamic networks, where the computational cost is adjusted per input sample at runtime \cite{yuan2023usdc, 9560049, Fernandez2025Cap}. A prominent example is early-exit mechanisms, which allow inference to terminate at intermediate layers based on input complexity and confidence levels, offering adaptive efficiency without altering the original backbone parameters, while only attaching auxiliary early-exit branches \cite{DBLP:journals/corr/abs-1810-07052}. This adaptivity is especially valuable in embedded scenarios, where computational effort can be dynamically matched to both input difficulty and available energy—conserving power during simple or idle periods and allocating full capacity only when necessary.

While static compression techniques like pruning and quantization have been extensively studied, and dynamic approaches such as early-exit networks represent a growing research direction, a critical gap remains. Existing literature typically benchmarks techniques within a single category—comparing different pruning methods or various quantization schemes—or focuses on improving a single technique in isolation. To the best of our knowledge, a unified empirical comparison across these categories, under consistent and realistic deployment conditions, is lacking. This leaves practitioners without clear guidance on selecting the most suitable optimization strategy for a given edge-AIoT scenario, particularly when considering novel dynamic paradigms like early exit.

In this study, we address this gap by providing a comprehensive comparison of prominent static compression (pruning, post-training quantization) and dynamic networks (early exit) techniques for \gls{cnn} inference on physical edge hardware. Rather than designing new architectures, we focus on applying these optimizations to existing models, evaluating both architectures specifically engineered for efficiency (e.g., MobileNetV2, ShuffleNet-V2, and EfficientNet-B2) and larger, more complex models (e.g., ResNet152).

\subsection{Main contributions and novelty}
Our evaluation framework is designed for  edge-\gls{aiot} scenarios and employs a widely-adopted inference engine to ensure realistic measurements (ONNX). We evaluate performance using a holistic set of metrics critical for edge deployment: accuracy, prediction stability metrics, latency (average and total), computational resource utilization (CPU and GPU usage), and model footprint (model size on disk and runtime RAM consumption). The main contributions of this work are:

\begin{itemize}
    \item A unified experimental framework providing a comprehensive comparison of static compression and dynamic networks techniques under identical conditions, using ONNX-based inference pipelines on representative edge devices to ensure fair and reproducible results.
    \item A comprehensive analysis of early exits as an edge optimization mechanism; we show they can achieve competitive accuracy–latency trade-offs and enable dynamic adaptability not available to static techniques. This empirical evidence highlights early exits as a practical and effective optimization strategy for \gls{aiot} deployments. 
    \item Real hardware benchmarking on heterogeneous platforms (such as NVIDIA Jetson Series) using a production-oriented inference engine (ONNX Runtime), reporting comprehensive metrics across a spectrum of CNNs architectures. This offers practitioners concrete guidance on selecting optimization methods based on deployment constraints, device capabilities, and operational requirements.
    \item A joint evaluation of static and dynamic optimization strategies, including their combined application (quantization with early exits), providing empirical insights into how these techniques interact in terms of accuracy, latency, and adaptability under edge constraints.
\end{itemize}

The remainder of this paper is organized as follows. Section \ref{sec:sta} provides the state of the art, detailing the optimization techniques and reviewing comparative studies and evaluation criteria relevant to our work. Section \ref{sec:methodology} describes our methodology, specifying the selected optimization techniques, evaluation metrics, and the experimental setup. Section \ref{sec:results} presents the results and analysis, beginning with an evaluation of accuracy and compression for each method, followed by an assessment of computational performance on edge devices, and concluding with a combination analysis of early exits and quantization strategies. Finally, Section \ref{sec:conclusions} concludes the paper with a summary of findings, practical implications, and suggestions for future research directions. Additional results and implementation details are provided in the Appendix \ref{app:pruning}.

\section{State of the art}\label{sec:sta}
\subsection{Model Optimization Techniques in \gls{ai}}
This section provides a detailed analysis of the main optimization techniques applied to \gls{cnn} together with early-exits mechanisms. These techniques address the challenge of deploying advanced models in environments with limited computational resources.

\subsubsection{Pruning}
Pruning is a widely used technique for reducing the complexity of \gls{dnn}, especially \gls{cnn}, removing or zeroing parameters (weights) or structures (neurons/filters) deemed low-importance \cite{mohanty2025pruning, Liu2017LearningEC}, thereby reducing the effective number of parameters, lowering memory consumption, and accelerating inference time, all while aiming to preserve accuracy \cite{Dantas_Comprehensive, Kuzmin_prunin}. As a result, deployment on resource-constrained devices becomes more efficient. In this context, pruning is typically categorized into two main methodologies: structured and unstructured.

Unstructured pruning is an optimization technique that ignores individual weights considered to be of low importance, explicitly assigning them a value of zero \cite{mohanty2025pruning}. These weights are not physically removed from the model; the corresponding connections still exist but no longer contribute to the computation. As a result, the total number of parameters does not decrease, and the network architecture remains unchanged. When the model is stored using sparse formats, both file size and storage costs can be reduced. This approach largely preserves model accuracy since it does not alter the overall architecture or eliminate entire structural units such as filters or channels. Nevertheless, the random distribution of zeroed weights leads to irregular sparsity patterns, which can hinder efficient execution on standard hardware \cite{wang2020sparsert, yao2019balanced, Liu2017LearningEC}, even when sparse memory formats are employed. Moreover, retraining or fine-tuning is typically required to compensate for potential accuracy degradation after pruning.

In the case of structured pruning, entire structures of the model, such as filters, channels, neurons, layers, or blocks, are permanently removed. This type of pruning produces a smaller network with a modified but completely dense architecture \cite{mohanty2025pruning}. This structural regularity makes it easier for the reduction in the number of parameters to translate directly into a decrease in inference time, memory usage, and computational cost \cite{Wang_Empowering, Liu2017LearningEC}. Therefore, structured pruning is particularly suitable for scenarios with limited resources. As with unstructured pruning, fine-tuning or additional training of the remaining network is typically required to compensate for any potential loss of accuracy.

In both methodologies, a criterion is required to determine which elements of the network should be pruned. A common approach is to use magnitude-based criteria, such as the L1-norm \cite{kumar2021pruning}. The L1-norm quantifies the overall magnitude of weights or filters and identifies those with the smallest values as less important, leading to a practical and hardware-agnostic criterion for guiding pruning decisions. Recent approaches, such as \cite{belhadi2025lightprune}, also rely on magnitude-based metrics while incorporating latency information to better adapt pruning to resource-constrained environments.

\subsubsection{Quantization}
Quantization is an optimization technique that reduces the size of models and speeds up inference by converting parameters (weights and activations) from high-precision formats (e.g., 32-bit floating point) to low-precision representations (8-bit or 4-bit integers) \cite{Kuzmin_prunin, Dantas_Comprehensive}. While this transformation decreases memory usage and computational requirements—facilitating deployment on resource-constrained edge devices—it may also introduce accuracy degradation, depending on the quantization scheme and target hardware.

There are different approaches to implementing quantization, including \gls{ptq}, \gls{dq}, and \gls{qat}. 
\gls{ptq} is a static quantization method applied after training. It uses a small calibration dataset to determine fixed scaling parameters for both weights and activations before deployment \cite{Agrawal_Efficient, Semerikov2025Edge}. Once calibrated, the entire model operates with lower precision during inference without further adjustments. In practice, static \gls{ptq} follows two main implementation patterns. The \textit{fused operator} approach replaces floating-point operations with quantized versions that embed scaling parameters directly, executing layers entirely in integer arithmetic \cite{KIM2025107718}. Alternatively, the \textit{insertion}-based strategy places explicit quantization and dequantization operations around each floating-point layer. While more flexible for hardware targeting, this pattern increases graph complexity.

\gls{dq} employs a hybrid scheme: weights are quantized statically before deployment (using predetermined scales), while activations are quantized dynamically at runtime based on their observed ranges \cite{Semerikov2025Edge}. This eliminates calibration data requirements but introduces per-inference scaling overhead. \gls{dq} is most beneficial for models where weight memory dominates, such as those with large linear or attention layers. For convolutional networks, however, the runtime cost of dynamic activation quantization often outweighs the benefits.

Finally, \gls{qat} incorporates simulated quantization within the model’s layers during training, enabling the network to learn to compensate for the effects of low-precision weights and activations. During the forward pass, weights and activations are quantized to int8 in a simulated manner, while the operations continue in float32 to allow gradient computation. During the backward pass, gradients are calculated normally in float32, allowing the model to adjust its parameters to mitigate quantization errors. After training, the model can be exported with actual int8 weights and activations, resulting in an efficient, low-precision model that maintains accuracy \cite{krishnamoorthi2018quantizing}. This approach generally provides the best trade-off between inference speed and performance, particularly in complex architectures where \gls{ptq} can cause substantial accuracy loss.

\subsubsection{Early Exits}
Early exits are a technique for accelerating inference in \gls{dnn}, particularly in \gls{cnn} models deployed on edge devices. The core idea is to introduce lightweight classifier branches at strategic layers of the network, enabling the model to generate intermediate predictions before reaching the final output layer \cite{Fernandez2025Cap}.

During inference, the network is traversed sequentially, layer by layer, until an early-exit branch is reached. At this point, the branch generates a prediction and its associated confidence is evaluated, typically using the softmax probability or the entropy $H(p)$ of the output distribution: 
\begin{equation}
    H(p) = -\sum_{i=1}^{n} p_i \log p_i,
\end{equation} where $p_i$ denotes the predicted probability for class $i$ among the $n$ possible classes. If the confidence exceeds a predefined \textit{\gls{threshold}}, the prediction is returned immediately, skipping the remaining layers and reducing computational cost and latency \cite{Fernandez2025Cap, teerapittayanon2016branchynet}.

The implementation typically involves three key steps: (1) identifying appropriate insertion points, often after major network blocks where features are sufficiently abstract; (2) designing the early-exit branches, which can range from simple classifier heads to more complex structures that include additional normalization or lightweight convolutional layers to further process intermediate features; and (3) choosing a training strategy, which can involve joint training with the base network, sequential training, or the use of weighted loss functions to balance the contribution of each branch. This flexible design enables input-dependent inference, where easy samples exit early while hard ones continue deeper into the network \cite{DBLP:journals/corr/abs-1810-07052}.

Although early exits do not alter the base architecture or reduce the number of parameters, they share with classical optimization techniques, such as pruning and quantization, the common goal of improving computational efficiency for deployment on resource-constrained devices. Unlike static methods, early exits offer dynamic inference paths that adapt to input complexity, making them a compelling alternative in edge scenarios where low latency and efficient resource use are critical.






\subsection{Comparative Studies and Evaluation Criteria}\label{sec:studies}
The optimization of \gls{cnn} for deployment on edge devices has traditionally been addressed using techniques such as pruning, and quantization. These strategies aim to reduce computational complexity, model size, inference latency, and resource consumption, thereby facilitating their execution in environments with limited capabilities.

However, existing comparative studies have significant limitations. One particularly notable aspect is that much research focuses exclusively on model performance metrics such as accuracy, precision, recall, and F1-score \cite{Kuzmin_prunin, Karathanasis2025, Naveen2024OptimizedCN, francy2024edgeaievaluationmodel}. Although in some cases, values related to latency or inference time are included, these are usually treated superficially, without in-depth analysis or a systematic approach. As a result, a complete view of the true impact these techniques have in edge scenarios, where computational efficiency and resource management are as important as predictive performance, is lost.

Methodologically, comparisons are often restricted to techniques within the same category—pruning versus pruning, or quantization versus quantization—rather than assessing which strategy is most effective for a given architecture or deployment scenario \cite{Li_model_compression, Karathanasis2025, Paranayapa2024}. While many studies focus on optimizing either large, over-parameterized networks (e.g., ResNet) or compact, edge-optimized models (e.g., MobileNet), to our knowledge, no study has systematically compared leading optimization paradigms across this entire architectural spectrum. Even in works that explore combinations—such as pruning with quantization or early exits with other techniques—a rigorous comparative framework is often lacking \cite{10909993, Korol_Pruning, saxena2023mcqueen, deutel2026recent}. These works typically present combined results without isolating the contribution of each technique or benchmarking against a comprehensive set of alternatives, particularly omitting cross-paradigm comparisons (e.g., static vs. dynamic). Consequently, it remains unclear whether static compression, dynamic adaptation, or their integration yields the best accuracy–latency trade-off for a specific edge target.

When inference performance is evaluated, a notable methodological gap persists. Comparative studies—even those restricted to a single optimization category like pruning or quantization—are frequently conducted on high-performance server GPUs rather than on actual edge hardware \cite{francy2024edgeaievaluationmodel, Karathanasis2025, Naveen2024OptimizedCN}. This practice obscures the true latency-accuracy trade-offs that arise from the interaction between dynamic runtime mechanisms and the specific memory, thermal, and power constraints of embedded systems. Although exceptions exist—such as PolyThrottle \cite{Yan_Polythrottle}, which uses a Jetson TX2, or SUQ \cite{Vaiyapuri2025}, tested on a mobile device—comprehensive comparisons that include dynamic techniques like early exits across diverse, real edge platforms remain scarce. Consequently, there is insufficient understanding of how these optimizations perform in the very environments where their efficiency gains are most critical.

In studies closer to the approach we propose, where early exits are included in the comparison, the work of \cite{Korol_Pruning} stands out. It implements a combined strategy of pruning and early exits. However, the study does not isolate the individual contribution of each technique, making it difficult to determine whether their combination provides synergistic benefits beyond their separate application. Furthermore, its experimental evaluation is confined to a single hardware platform (a \gls{fpga}) and does not explore other optimization techniques, such as quantization. This narrow scope leaves open important questions about the generalizability of the findings across diverse edge devices and broader optimization pipelines.

Alternatively, the work presented in \cite{Xu_EfficientHardware} explores the relationship between quantization and early exits, also in the context of an \gls{fpga} as an edge platform. However, the analysis is limited to presenting the results of each technique separately, without carrying out a direct and systematic comparison between the two. This makes it difficult to draw solid conclusions about their relative effectiveness or their possible complementarity in real scenarios.

Apart from early exits, other dynamic network techniques exist that aim to reduce computation, such as layer skipping, which dynamically skips internal layers based on input difficulty \cite{Wang_2018_ECCV}, and token pruning, which discards less informative tokens (e.g., redundant words or image patches) \cite{10.5555/3540261.3541329}. While these methods adapt computation to the input, they often involve more complex control logic and can degrade accuracy. In contrast, early exits are easier to integrate into existing architectures and typically offer a better trade-off between accuracy and efficiency, making them particularly attractive for edge deployment.

To address the mentioned gaps, this work presents a systematic hardware-in-the-loop comparison of the dominant edge optimization paradigms: pruning (static), quantization (static), and early exits (dynamic). We evaluate them individually and in combination across multiple edge devices and a diverse set of architectures—from large, over-parameterized networks to compact, edge-optimized models—measuring accuracy, inference latency, model size, and memory footprint. 

\section{Methodology}\label{sec:methodology}



Three model optimization techniques are evaluated: pruning, quantization, and early exits. The techniques are applied to \gls{cnn} models deployed on real edge devices. The full-precision (FP32) model for each architecture serves as the baseline against which all optimized variants are compared.

To ensure a representative and diverse benchmark, four architectures are selected: ResNet-152, EfficientNet-B2, MobileNet-V2, and ShuffleNet-V2. ResNet-152 represents a large, high-capacity model, allowing us to analyze how optimization techniques affect deep architectures with substantial computational demands. Conversely, MobileNet-V2 and ShuffleNet-V2 are lightweight architectures designed explicitly for edge environments, enabling an assessment of optimization impacts on already parameter-efficient models. EfficientNet-B2 provides a balanced intermediate point. 

All models are trained on ImageNet100, exported to the standardized ONNX format, and evaluated using ONNX Runtime. This pipeline ensures reproducibility, cross-platform compatibility, and alignment with real-world edge deployment scenarios where lightweight, portable inference runtimes are essential.

\subsection{Optimization Techniques}\label{sec:methodology_opt}
We apply and evaluate three model optimization techniques: quantization, pruning, and early-exit mechanisms. 

\subsubsection{Pruning}
We apply structured pruning to reduce both model size and inference time. Pruning is performed on the pre-trained base model, permanently modifying its architecture. To enhance execution efficiency on GPU-based platforms, channel rounding aligns the number of channels to fixed granularities (multiples of 16 or 32), thereby exploiting hardware-level optimizations like memory alignment and block-level parallelism \cite{9042000, LIU2021100009}.

However, in the case of ShuffleNet-V2, this pruning methodology is intrinsically incompatible with the network architecture due to its channel shuffle operation \cite{zhao2024streamlining, zhao2022probability}. This operation establishes a rigid, inter-dependent mapping between channel groups through tensor reshaping and transposition. Removing arbitrary channels via standard pruning breaks this mapping, leading to tensor dimension mismatches and execution failures. While specialized pruning strategies exist to handle this constraint \cite{Gong2023Dynamic}, we deliberately exclude ShuffleNet-V2 from the pruning experiments to maintain a consistent and comparable experimental framework for all other models.

We evaluate multiple pruning configurations by varying both the \gls{pr} (proportion of filters removed) and the \gls{cg} (minimum number of channels pruned per step). These parameters allow us to analyze the impact of pruning aggressiveness on model size, computational cost, and predictive accuracy. All configurations and results are detailed in Appendix \ref{app:pruning}. The configuration offering the best trade-off between accuracy and model size reduction is selected for the main experiments, as a smaller model size directly translates to lower inference latency on edge hardware.

\subsubsection{Quantization}\label{sec:methodology_quant}
We evaluate \gls{dq} and \gls{ptq}, targeting unsigned 8-bit integer (UINT8) precision for weights and activations. These techniques are widely adopted in edge deployment due to their applicability to pre-trained models without architectural changes or retraining, ensuring fair and consistent comparisons across optimization techniques in our study. 

A key advantage of \gls{dq} and \gls{ptq} is their seamless compatibility with standard ONNX-based deployment pipelines. Quantization can be applied directly to exported ONNX models, enabling straightforward integration with ONNX Runtime and edge-oriented inference toolchains—a critical requirement for our experimentally realistic setup.

In the case of \gls{ptq}, we employ a layer-wise selective quantization strategy to maximize accuracy, particularly in architectures known to be sensitive to uniform quantization. Based on a preliminary sensitivity analysis, we explicitly configure which operation types to quantize, excluding those that cause significant accuracy degradation. For EfficientNet-B2, we quantize only convolutional layers (Conv), leaving all other operations in FP32. For ShuffleNetV2, we quantize convolutional, fully connected, and matrix multiplication layers (Conv, Gemm, MatMul). For quantization-friendly architectures such as ResNet-152 and MobileNet-V2, we apply quantization to the entire computational graph. Based on this analysis, we categorize architectures as either quantization-friendly—those where full-graph quantization preserves accuracy—or quantization-sensitive—those requiring selective layer exclusion to avoid severe degradation. This fine-grained control, which allows us to isolate and protect sensitive subgraphs, is a unique capability enabled by PTQ's static graph analysis.

We deliberately exclude \gls{qat} to maintain a unified and easily deployable pipeline across all architectures and optimization strategies. While \gls{qat} can achieve higher accuracy by incorporating quantization effects during training \cite{nagel2021white, krishnamoorthi2018quantizing}, it introduces significant complexity in the model export and deployment process. In practice, \gls{qat} relies on fake-quantization operators and framework-specific constructs that do not always map cleanly to standardized quantized ONNX operators. 
Supporting \gls{qat} would therefore require custom operator conversions or specialized quantization frameworks, such as Brevitas \cite{brevitas2021}, which are tightly coupled to specific toolchains and often lack direct compatibility with standard ONNX Runtime workflows. Since our objective is to evaluate optimization techniques under realistic deployment constraints, we restrict our analysis to \gls{dq} and \gls{ptq}, which provide a favorable balance between accuracy, implementation complexity, and deployability on real edge hardware.

\subsubsection{Early Exits}
The training of the early exits follows the BranchyNet framework \cite{teerapittayanon2016branchynet} in its use of auxiliary cross-entropy losses for each exit, encouraging early and accurate predictions. However, unlike the original end-to-end training approach, we adopt a two-stage/classifier-wise training \cite{Fernandez2025Cap, kubaty2025how, bakhtiarnia2021multi} strategy: the backbone network is kept frozen, and only the parameters of the early-exit branches are trained. This design choice ensures a fair comparison across optimization techniques by preserving the representational capacity of the pretrained model \cite{kubaty2025how} while also preventing interference between branches during training. Crucially, by freezing the backbone, the gradients from each exit only update its own dedicated branch parameters. This prevents interference between exits during simultaneous training and maintains the stability of the original feature extractor.

Regarding the design of the early-exit branches, different architectures require different levels of complexity. In the case of ResNet-152, a large high-capacity model, each early exit consists of a simple classifier composed of pooling and flattening layers followed by a fully connected layer. This design is sufficient to effectively leverage the rich intermediate feature representations produced by the network while ensuring architectural consistency across exit points. In contrast, for EfficientNet-B2, MobileNet-V2, and ShuffleNet-V2, the early exits require a more elaborate structure. Due to their highly optimized and parameter-efficient nature, these lightweight architectures are unable to learn sufficiently discriminative representations at early layers using a simple classifier alone \cite{10.1145/3587135.3592204}. Prior work has shown that incorporating one of the internal blocks of the network into the early-exit branch provides a better balance between predictive capability and architectural efficiency, and this strategy is therefore adopted for these models.

Although all early exits are trained simultaneously in a single forward/backward pass, three separate inference models are derived to analyze the impact of exit location on performance. Each inference model contains only one early exit placed at a specific point in the network, together with the corresponding truncated backbone. These models are denoted as ``EEi'', where i indicates the sequential position (by depth) within the original network where an early exit is possible and has been inserted. For example, in ResNet-152, EE2, EE8, and EE14 correspond to exits placed at the second, eighth, and fourteenth possible insertion points, respectively, representing locations near the early, intermediate, and deeper stages of the network. This setup allows us to isolate the effect of exit placement while avoiding confounding interactions between multiple exits during inference.

During inference, each early-exit model is evaluated under different operational conditions by sweeping a range of confidence \textit{\gls{threshold}}s. The confidence \textit{\gls{threshold}} determines whether a sample exits at the early branch or continues to the final classifier. Lower \textit{\gls{threshold}}s increase the proportion of samples exiting early, reducing inference latency and computational cost at the potential expense of accuracy, whereas higher \textit{\gls{threshold}}s result in fewer early exits and higher accuracy but increased computational overhead. From this sweep, two representative operating points are selected for each exit: \textit{acc-opt}, which maximizes overall accuracy, and \textit{inf-opt}, which prioritizes inference efficiency and resource reduction. This methodology enables a systematic analysis of how early-exit location and confidence \textit{\gls{threshold}} jointly affect accuracy, latency, and computational cost across different architectures and optimization strategies.


\subsection{Experimental setup}\label{sec:methodology_exp}
All models were deployed using ONNX Runtime (version 1.19.0), a cross-platform inference engine designed to efficiently run models in ONNX format. While this framework ensures compatibility with a wide variety of hardware and software environments, incorporating early-exit mechanisms introduces non-trivial challenges when exporting models. In practice, although ONNX provides limited support for control-flow operators, complex early-exit policies implemented in PyTorch, such as conditional execution based on confidence thresholds using operators like \textit{torch.cond}, cannot be reliably translated during export \cite{pytorch_torch_cond}.

The core issue is a representational mismatch. There is no direct and stable mapping between PyTorch’s dynamic conditional constructs and static ONNX operators, especially when branches contain entire sub-networks. This reflects a broader difficulty in model conversion: complex, dynamic control flow is inherently challenging to capture in static graph formats like ONNX, often leading to conversion failures or semantically incorrect models \cite{10.1145/3650212.3680374}. Consequently, an early-exit architecture with integrated stopping criteria cannot be represented as a single ONNX graph. 

To address this restriction, a strategy based on model partitioning was adopted: Instead of exporting a single graph with conditional logic, the entire model is divided into a sequence of independent ONNX submodels. Each submodel corresponds to a specific early-exit point and processes the input only up to that block. After each submodel is executed, a confidence criterion is evaluated; if the prediction exceeds the set \textit{\gls{threshold}}, the inference ends, otherwise, the generated output is used as input for the next submodel in the sequence. In this way, the system emulates a dynamic control flow using only the standard sequential execution provided by ONNX Runtime \cite{XU2025122470}.

Building on this ONNX structure, we optimize inference by selecting platform-specific execution providers: CUDA for NVIDIA Jetson devices (leveraging GPU acceleration) and CPU-only execution for the Raspberry Pi 5 (which lacks a dedicated GPU). Jetson devices employ a unified memory architecture in which the GPU and CPU share the same physical RAM. During GPU-accelerated inference, model parameters, CUDA contexts, and intermediate tensors reside in main memory, significantly increasing RAM usage compared to CPU-only execution. This feature directly impacts resource trade-offs in embedded edge computing systems \cite{Piveral_Memory}.

The environment setup ensures reproducibility and efficiency: JetPack 6.0 with Python 3.11 is used on the Jetson devices, while a native CPU-only environment is run on the Raspberry Pi 5. The batch size is set to 1 for all tests, simulating real-time inference scenarios on edge. The hardware specifications are shown in Table \ref{tab:devices}.

To isolate the effects of the model-level optimizations, we deliberately avoid using highly specialized inference engines like NVIDIA TensorRT. Such engines apply aggressive, proprietary kernel-level optimizations that would obscure the direct impact of pruning, quantization, and early exits on performance. Our choice of ONNX Runtime provides a more neutral and consistent baseline, ensuring that the reported improvements or trade-offs are a direct consequence of the applied optimization techniques.

\begin{table}[htbp]
\caption{Edge Device Specifications}
\label{tab:devices}
\begin{center}
\begin{tabular}{|c|c|c|c|}
\hline 
\textbf{Device} & \textbf{Memory} & \textbf{GPU} & \textbf{CPU} \\ 
\hline 
\hline
\makecell{Jetson \\ AGX Orin} & \makecell{64GB \\ LPDDR5} & 2048-core & \makecell{Arm Cortex-A78AE \\ 12-cores (2.2\,GHz)} \\    
\hline
\makecell{Jetson \\ Orin Nano} & \makecell{8GB \\LPDDR5} & 1024-core & \makecell{Arm Cortex-A78AE \\ 6-cores (1.5\,GHz)}\\  
\hline
\makecell{Raspberry \\ Pi 5} & \makecell{8GB \\ LPDDR4x} & N/A & \makecell{BCM2712 Arm Cortex-A76 \\ 4-cores (2.4\,GHz)}\\
\hline 
\end{tabular}
\label{tab2}
\end{center}
\end{table}

\subsection{Evaluation Metrics}\label{sec:evalu_metrics}
To analyze the trade-offs introduced by each optimization technique, we adopt a set of evaluation metrics that capture both model behavior and hardware-related effects. The metrics are grouped into two complementary aspects: model performance and system resource utilization.

In terms of model performance, indicators that measure compression, inference behavior, and predictive quality are taken into consideration \cite{Agrawal_Efficient}. The overall \textit{accuracy} serves as the primary metric for evaluating the classification capability of the model after optimization, ensuring that compression does not critically undermine its core function. The \textit{compression ratio} reflects the ratio between the original and the optimized model size, providing a direct measure of memory savings. The \textit{total inference time}, measured in seconds, quantifies how long it takes the model to process the entire dataset, including image preprocessing. Meanwhile, the \textit{average inference time}, expressed in milliseconds, represents the mean time required to process a single input sample, averaged over the evaluation set. 

In models that incorporate early exits, additional measurements are considered to characterize the internal behavior of the early-exit mechanism during inference. The \textit{early-exit rate} measures the percentage of input samples that are resolved at an early exit, directly quantifying the effectiveness of the dynamic networks mechanism. Consequently, the \textit{average inference time} is reported separately for samples exiting at early branches and for samples reaching the final classifier. This distinction allows for a more accurate understanding of the latency benefits of dynamic networks based on confidence.

In addition to direct performance, metrics have been included to evaluate the consistency and stability of the model's decisions after optimization. \textit{Label loyalty} indicates the percentage of agreement between the labels predicted by the optimized model and those of the original model. This metric provides a clear view of the extent to which classification decisions are preserved. Complementarily, \textit{probability loyalty} measures the similarity between the output probability distributions of the optimized and baseline models, capturing shifts in predictive confidence beyond mere accuracy \cite{Dantas_Comprehensive}.

On the other hand, \textit{system resource} usage during inference is analyzed to assess deployment feasibility on devices with limited resources. CPU usage is measured as the average load across all logical cores using psutil~\cite{psutil_cpu_percent}. GPU usage on CUDA-enabled platforms, such as Jetson devices, is monitored with jetson-stats (jtop) \cite{jetsonstats_gpu_class} in terms of load percentage and memory usage. RAM usage of the inference process is recorded using psutil’s Resident Set Size, which reflects the actual physical memory occupied by the process. Metrics such as Virtual Memory Size (vms) or shared either include memory not resident in RAM or memory that can be shared with other processes, making RSS the most accurate indicator of the memory footprint required for inference \cite{jetsonstats_jtop_memory}.


\section{Results and analysis}\label{sec:results}
This section provides an in-depth analysis of the implementation of pruning, quantization, and early-exit strategies within the context of the ResNet-152, EfficientNet-B2, MobileNet-V2, and ShuffleNet-V2 architectures, specifically for edge deployment scenarios. We structure the analysis in three parts: first, the predictive performance metrics; second, its real-world inference performance on the target hardware; and third, the synergistic potential of combining multiple techniques.

\subsection{Evaluation of the accuracy and compression of edge-optimized models}\label{sec:results_A}
This first analysis focuses on the intrinsic properties of the optimized models, independent of deployment hardware. We evaluate the impact of each optimization technique on predictive accuracy, compression ratio, and prediction consistency, the latter quantified through label loyalty and probability loyalty metrics as defined in Section~\ref{sec:evalu_metrics}.

Table~\ref{tab:accuracy_model_metrics} summarizes the results across all architectures and configurations, reporting model size (MB), compression ratio (relative to the baseline), accuracy (\%), label loyalty (\%), and probability loyalty (\%). For models incorporating early exits, the table additionally reports the early-exit rate (\%), which represents the proportion of samples that exit through an early-exit branch rather than the final-exit layer. The evaluated configurations include the base model (BASE), structured pruning (PRUNE) as selected in Appendix~\ref{app:pruning}, 8-bit quantized models using both \gls{ptq} and \gls{dq}, and multiple early-exit (EE) configurations. 

\begin{table*}[ht]
\caption{model fidelity and compression metrics}
\label{tab:accuracy_model_metrics}
\begin{center}
\tiny
\begin{tabular}{|c|c|r|r|r|r|r|r|}
\hline
\textbf{Architecture} & \textbf{Model} &  \makecell{\textbf{Size} \\ \textbf{(MB)}} & \textbf{Compression} & \makecell{\textbf{Accuracy} \\ \textbf{(\%)}} & \makecell{\textbf{Label Loyalty} \\ \textbf{(\%)}} & \makecell{\textbf{Prob. Loyalty} \\ \textbf{(\%)}} & \makecell{\textbf{Early-Exit} \\ \textbf{Rate (\%)}} \\ 
\hline
\hline 
\multirow{10}{*}{\textbf{ResNet152}} & \textbf{BASE} & \textbf{233.20} & \textbf{1.00x} & \textbf{89.08} & - & - & - \\
\cline{2-8}
& PRUNE & 98.50 & 2.37x & 87.22 & 90.88 & 94.11 & -  \\
\cline{2-8}
& PTQ & \textcolor{good}{59.10} & \textcolor{good}{3.95x} & 84.30 & 88.32 & 91.95 & - \\
& DQ & 59.20 & 3.94x & 88.96 & 97.96 & 97.37 & - \\
\cline{2-8}
& EE3 (\textit{acc-opt}, T = 0.01) & \multirow{2}{*}{\textcolor{bad}{233.60}} & \multirow{2}{*}{\textcolor{bad}{0.99x}} & \textcolor{good}{89.08} & \textcolor{good}{99.98} & \textcolor{good}{99.98} &  5.66 \\
& EE3 (\textit{inf-opt}, T = 0.54) &  &  & \textcolor{bad}{70.86} & 75.76 & 79.32 & 79.30 \\
\cline{2-8}
& EE8 (\textit{acc-opt}, T = 0.019) & \multirow{2}{*}{\textcolor{bad}{233.60}} & \multirow{2}{*}{\textcolor{bad}{0.99x}} & \textcolor{good}{89.08} & \textcolor{good}{99.98} & \textcolor{good}{99.98} & 11.34 \\
& EE8 (\textit{inf-opt}, T = 0.55) &  &  & 71.08 & \textcolor{bad}{75.12} & \textcolor{bad}{79.51} & 88.28 \\
\cline{2-8}
& EE14 (\textit{acc-opt}, T = 0.018) & \multirow{2}{*}{\textcolor{bad}{233.60}} & \multirow{2}{*}{\textcolor{good}{0.99x}} & 89.06 & 99.86 & 99.87 & 25.66 \\
& EE14 (\textit{inf-opt}, T = 0.65) &  &  & 74.84 & 76.54 & 81.50 & 99.62 \\
\hline
\hline
\multirow{10}{*}{\textbf{EfficientNet-B2}} & \textbf{BASE} & \textbf{31.30} & \textbf{1.00x} & \textbf{91.18} & - & - & - \\
\cline{2-8}
& PRUNE & 25.60 & 1.20x & 87.12 & 89.48 & 92.71 & - \\
\cline{2-8}
& PTQ & 9.00 & 3.47x & 64.86 & 66.06 & 71.64 & - \\
& DQ & \textcolor{good}{8.40} & \textcolor{good}{3.73x} & \textcolor{bad}{11.08} & \textcolor{bad}{11.00} & \textcolor{bad}{15.47} & - \\
\cline{2-8}
& EE3 (\textit{acc-opt}, T = 1e-6)& \multirow{2}{*}{31.40} & \multirow{2}{*}{0.99x} & 91.10 & 99.92 & 99.92 & 0.08 \\
& EE3 (\textit{inf-opt}, T = 0.45) &  &  & 38.86 & 42.00 & 42.60 & 60.48 \\
\cline{2-8}
& EE8 (\textit{acc-opt}, T = 0.005) & \multirow{2}{*}{\textcolor{bad}{31.53}} &\multirow{2}{*}{\textcolor{bad}{0.99x}} & \textcolor{good}{91.20} & \textcolor{good}{99.98} & \textcolor{good}{99.98} & 7.42 \\
& EE8 (\textit{inf-opt}, T = 0.45) &  &  & 77.78 & 82.52 & 85.50 & 75.74  \\
\cline{2-8}
& EE15 (\textit{acc-opt}, T = 0.01) & \multirow{2}{*}{32.30} & \multirow{2}{*}{0.97x} & 91.18 & 99.66 & 99.68 & 50.12 \\
& EE15 (\textit{inf-opt}, T = 0.45) &  &  & 83.38 & 84.62 & 87.87 & 98.64 \\
\hline
\hline
\multirow{10}{*}{\textbf{MobileNet-V2}} & \textbf{BASE} & \textbf{9.40} & \textbf{1.00x} & \textbf{87.80} & - & - & - \\
\cline{2-8}
& PRUNE & 6.80 & 1.38x & 78.82 & 81.76 & 87.41 & - \\
\cline{2-8}
& PTQ & 2.60 & 3.62x & 81.22 & 85.88 & 90.92 & - \\
& DQ & \textcolor{good}{2.50} & \textcolor{good}{3.76x} & 86.82 & 95.02 & 98.27 & - \\
\cline{2-8}
& EE3 (\textit{acc-opt}, T = 0.07)& \multirow{2}{*}{9.47} & \multirow{2}{*}{0.99x} & 87.84 & \textcolor{good}{97.58} & \textcolor{good}{97.56} & 0.50   \\
& EE3 (\textit{inf-opt}, T = 0.6) &  &  & \textcolor{bad}{51.60} & \textcolor{bad}{57.14} & \textcolor{bad}{62.29} & 54.62 \\
\cline{2-8}
& EE8 (\textit{acc-opt}, T = 0.13) & \multirow{2}{*}{9.58} &\multirow{2}{*}{0.98x} & \textcolor{good}{87.90} & 97.56 & 99.38 & 8.58 \\
& EE8 (\textit{inf-opt}, T = 0.67) &  &  & 70.10 & 75.48 & 80.25 & 73.64 \\
\cline{2-8}
& EE13 (\textit{acc-opt}, T = 0.1) & \multirow{2}{*}{\textcolor{bad}{9.70}} & \multirow{2}{*}{\textcolor{bad}{0.97x}} & 87.86 & 97.56 & 99.41 & 16.32 \\
& EE13 (\textit{inf-opt}, T = 0.85) &  &  & 74.34 & 76.24 & 81.45 & 99.18 \\
\hline
\hline
\multirow{10}{*}{\textbf{ShuffleNet-V2}} & \textbf{BASE} & \textbf{5.50} & \textbf{1.00x} & \textbf{77.86} & - & - & - \\
\cline{2-8}
& PRUNE & - & - & - & - & - & - \\
\cline{2-8}
& PTQ & \textcolor{good}{1.50} & \textcolor{good}{3.67x} & 76.84 & 95.08 & 99.20 & - \\
& DQ & \textcolor{good}{1.50} & \textcolor{good}{3.67x} & 75.74 & 89.34 & 96.82 & - \\
\cline{2-8}
& EE2 (\textit{acc-opt}, T = 1e-7) & \multirow{2}{*}{5.70} & \multirow{2}{*}{0.96x} & 77.80 & \textcolor{good}{99.94} & \textcolor{good}{99.94} & 0.10 \\
& EE2 (\textit{inf-opt}, T = 15e-4 ) &  &  & \textcolor{red}{40.34} & \textcolor{bad}{50.10} & \textcolor{bad}{50.47} & 0.51  \\
\cline{2-8}
& EE5 (\textit{acc-opt}, T = 0.03) & \multirow{2}{*}{5.88} & \multirow{2}{*}{0.94x} & 77.90 & 99.22 & 99.37  & 20.38 \\
& EE5 (\textit{inf-opt}, T = 0.4) &  &  & 73.24 & 86.62 & 90.22 & 61.94 \\
\cline{2-8}
& EE8 (\textit{acc-opt}, T = 0.1 ) & \multirow{2}{*}{\textcolor{bad}{5.90}} & \multirow{2}{*}{\textcolor{bad}{0.93x}} & \textcolor{good}{78.18} & 98.20 &  98.41 & 36.92 \\
& EE8 (\textit{inf-opt}, T = 0.6) &  &  & 73.72 & 81.70 & 87.35 & 73.76 \\
\hline
\end{tabular}
\end{center}
\end{table*}

To facilitate comparison, the baseline model is shown in bold to serve as a reference point, while the best and worst values for each metric are highlighted in green and red, respectively, indicating the most and least favorable outcomes among the evaluated techniques. 

\textbf{Pruning} demonstrates a strongly architecture-dependent behavior. In large, over-parameterized models such as ResNet-152, structured pruning achieves a significant compression ratio (2.37x) while maintaining a relatively modest accuracy drop of -1.86\% (from 89.08\% to 87.22\%). High label and probability loyalty (90.88\% and 94.11\%, respectively) further indicate that the pruned model preserves the original decision structure for most samples. This confirms that deep, high-capacity architectures contain significant parameter redundancy that can be removed with limited impact on predictive performance.

In contrast, the effectiveness of pruning degrades markedly for architectures explicitly designed for efficiency, such as EfficientNet-B2 and MobileNet-V2. For these models, pruning yields only marginal compression gains (1.20x and 1.38x, respectively) while inducing substantially larger accuracy drops (-4.06\% for EfficientNet-B2 and -8.98\% for MobileNet-V2). This behavior reflects the fact that edge-optimized architectures already employ aggressive parameter sharing, depthwise separable convolutions, and compound scaling strategies, leaving limited redundancy to exploit. Consequently, as discussed in Appendix~\ref{app:pruning}, achieving an acceptable accuracy–compression trade-off in these models requires significantly lower pruning ratios, which directly limits the compression and speed-up that can be attained. As a result, pruning tends to remove functionally critical weights, leading to disproportionate performance degradation relative to the achieved compression.

The architectural dependency is further underscored by the case of ShuffleNet-V2, which was excluded from the pruning analysis due to fundamental incompatibilities with our standard methodology, as detailed in Section~\ref{sec:methodology_opt}

Overall, these results highlight that pruning is most effective when applied to large, generic backbones, whereas its benefits diminish significantly for compact architectures where efficiency is already baked into the design. This observation motivates the exploration of complementary optimization strategies—such as quantization and early-exit mechanisms—for edge-oriented models, where further structural compression alone may be insufficient.

\textbf{Quantization} provides consistent and substantial model compression, offering the best overall trade-off between model size and accuracy, with an average 3.5x-3.7x reduction in size across all architectures. However, the optimal quantization strategy depends critically on the model's architectural sensitivity. Our results reveal a clear pattern: \gls{dq} excels in  inherently quantization-friendly architectures architectures such as ResNet-152 and MobileNet-V2. For these models, where the primary challenge is precise activation range calibration, \gls{dq}'s runtime adaptation avoids errors introduced when \gls{ptq} must estimate fixed scales from limited calibration data. This allows \gls{dq} to closely match the full-precision baseline (e.g., ResNet-152 \gls{dq}: 88.96\% vs. FP32: 89.08\%; MobileNet-V2 \gls{dq}: 86.82\% vs. FP32: 87.80\%) and, crucially, in model fidelity: label loyalty exceeds 95\% and probability loyalty 97\% in both cases, confirming that \gls{dq} preserves not only accuracy but also the original model's decision behavior. 

Conversely, \gls{ptq} achieves higher accuracy in architectures containing operations susceptible to quantization noise, namely EfficientNet-B2 and ShuffleNet-V2. The superiority of \gls{ptq} in these cases is directly due to the layer-wise, selective quantization strategy detailed in Section~\ref{sec:methodology_quant}. For ShuffleNet-V2, applying this selective approach (quantizing only Conv, Gemm, and MatMul layers) allows \gls{ptq} (76.84\%) to outperform \gls{dq} (75.74\%) while achieving identical 3.67x compression.

This architectural sensitivity is most pronounced in EfficientNet-B2, where \gls{dq} results in catastrophic accuracy degradation (11.08\%), with label loyalty and probability loyalty collapsing to just 11.00\% and 15.47\%, respectively. In contrast, \gls{ptq} with selective quantization (applied only to convolutional layers) achieves substantially higher accuracy (64.86\%). This dramatic 53.8\% accuracy drop in \gls{dq} stems from the intrinsic vulnerability of EfficientNet-B2 architectures to indiscriminate quantization, primarily because they use depthwise separable convolutions and smooth non-linear activation functions such as Swish (SiLU) \cite{DBLP:journals/corr/abs-2004-09576}. These components generate highly asymmetric activation distributions characterized by outliers, resulting in extended dynamic ranges. While \gls{ptq}'s static graph analysis can mitigate this by isolating sensitive layers with complete precision, \gls{dq}'s uniform runtime-scale estimation causes quantization noise to propagate and amplify catastrophically throughout the model.

In summary, quantization delivers consistent compression (3.47x–3.95x). However, with varying trade-offs in accuracy, \gls{dq} provides near-lossless compression for quantization-friendly architectures such as ResNet-152 and MobileNet-V2. At the same time, \gls{ptq} with selective layer quantization is essential for maintaining usable accuracy in sensitive architectures like EfficientNet-B2 and ShuffleNet-V2. The adaptive nature of \gls{dq} is insufficient when specific architectural components act as critical accuracy bottlenecks, indicating that effective compression for such models requires either fine-grained PTQ control or training-aware approaches such as \gls{qat} (discarded in our study as discussed in Section~\ref{sec:methodology_quant}).

\textbf{Early Exit (EE)} configurations clearly illustrate the trade-off between accuracy and early-exit rate — the proportion of samples that exit before reaching the final layer — while introducing a modest parameter overhead from the additional classification heads, visible as a slight increase in model size (e.g., ShuffleNet-V2 EE2: 5.70 MB vs. BASE: 5.50 MB). As designed, \textit{acc-opt} preserve the reference model's accuracy with near-perfect fidelity — often within ±0.1\% — but yield very low early-exit rates, offering little computational advantage despite the modest increase in model size. For instance, ResNet-152 EE14 \textit{acc-opt} achieves 89.06\% (vs. BASE: 89.08\%), and EfficientNet-B2 EE8 \textit{acc-opt} reaches 91.20\% (vs. BASE: 91.18\%). In some configurations, the added classifiers even slightly improve the baseline accuracy, as seen in ShuffleNet-V2 EE8 \textit{acc-opt} (78.18\% vs. BASE: 77.86\%) and MobileNet-V2 EE8 \textit{acc-opt} (87.90\% vs. BASE: 87.80\%). This indicates that, for certain samples where intermediate layers produce sufficiently clear features, the additional early-exit classifiers can correctly predict the class even when the baseline might fail, as observed in ShuffleNet-V2 EE8 (78.18\% vs. BASE: 77.86\%) and MobileNet-V2 EE8 (87.90\% vs. BASE: 87.80\%).

Conversely, \textit{inf-opt} force high early-exit rates to reduce computational cost, incurring substantial but variable accuracy penalties. In ResNet-152, deeper early-exits (EE14 \textit{inf-opt}: 74.84\%) recover some accuracy compared to shallower ones (EE3 \textit{inf-opt}: 70.86\%). However, the shallower early-exit maximizes potential computational savings per exited sample (achieving an exit rate of 79.3\%). For the lightweight MobileNet-V2 and ShuffleNet-V2, the most aggressive shallow exits result in severe accuracy degradation (e.g., MobileNet-V2 EE3 \textit{inf-opt}: 51.60\%, a 36.2\% drop; ShuffleNet-V2 EE2 \textit{inf-opt}: 40.34\%, a 37.5\% drop). In this context, intermediate early exits provide a more viable \textit{inf-opt} operating point, achieving better accuracy (e.g., ShuffleNet-V2 EE5 \textit{inf-opt}: 73.24\%) while allowing significant computation skipping (61.94\% early-exit rate). This observation indicates that in compact architectures, the very early features lack sufficient discriminative information, suggesting that an intermediate network depth is the optimal compromise for latency-sensitive deployments.

On the other hand, the search for optimal thresholds highlights their high architecture and exit-specific sensitivity, with values varying across several orders of magnitude (e.g., ShuffleNet-V2 EE2 \textit{acc-opt}: T=1e-7 vs. MobileNet-V2 EE13 \textit{inf-opt}: T=0.85). This illustrates the necessity for independent calibration at each early exit to balance its distinct confidence distribution with the desired operating point for accuracy-efficiency.

Overall, early exits offer a dynamic trade-off between early-exit rate and accuracy that depends on two key factors: exit depth and confidence threshold. \textit{Acc-opt} maintain fidelity close to the reference value, demonstrating that the mechanism can be nearly lossless (e.g., ResNet-152 EE14 \textit{acc-opt}: 89.06\% vs. BASE: 89.08\%). In contrast, \textit{inf-opt} configurations show that it is possible to exit most samples early — over 99\% in ResNet-152 EE14 and MobileNet-V2 EE13 — but at the cost of considerable architecture-dependent accuracy loss. The technique proves most effective when intermediate features are sufficiently informative to enable early, reliable decisions. Large architectures such as ResNet-152 better absorb this trade-off, retaining over 74\% accuracy even at exit rates above 99\%, while lighter models suffer sharper degradation. This highlights the importance of meticulous exit placement and threshold calibration to ensure successful implementation.

\textbf{In summary}, the results show that quantization is the most effective technique for balancing compression and accuracy, reducing size by 3.5x–4x while maintaining accuracy close to the reference value across three of the four architectures. Pruning offers reliable but modest gains (between 1.2x and 2.4x compression) and serves as an alternative for quantization-sensitive models such as EfficientNet-B2. Early exits, while introducing a slight parameter overhead, achieve the highest accuracy among all optimization methods — often matching or even slightly exceeding the baseline, demonstrating that the mechanism can be lossless when optimized for fidelity. However, their \textit{inf-opt} modes incur substantial accuracy penalties despite reaching early-exit rates above 99\%.

Architecture-specific results reveal clear avenues for optimization: ResNet-152 excels at both quantization and pruning; EfficientNet-B2 is limited to pruning; and MobileNet-V2 and ShuffleNet-V2 achieve optimal results through quantization (\gls{dq} and \gls{ptq}, respectively). Early exits, meanwhile, offer a complementary dimension: they enable dynamic inference with minimal accuracy loss when configured appropriately, but their benefits — reduced computational cost and latency — are not reflected in static metrics such as model size. These advantages require task-specific latency measurements to be adequately assessed. This comparative analysis confirms that, although quantization tends to predominate in static compression, architectural limitations dictate when pruning is necessary and in which cases early exits can provide runtime flexibility.

\subsection{Evaluating the inference performance of optimization methods on edge devices}\label{sec:results_B}

\begin{table*}[t]
\caption{Edge device performance metrics utilization ResNet-152}
\label{tab:performance_resnet}
\begin{center}
\tiny
\begin{tabular}{|c|c|c@{\hspace{1mm}}c|c@{\hspace{1mm}}c|c@{\hspace{1mm}}c|c@{\hspace{1mm}}c|c@{\hspace{1mm}}c|c@{\hspace{1mm}}c|c@{\hspace{1mm}}c|}
\hline
\multirow{2}{*}{\textbf{Device}} & \multirow{2}{*}{\textbf{Model}}
    & \multicolumn{2}{c|}{\makecell{\textbf{Total Inf.} \\ \textbf{Time (s)}}}  
    & \multicolumn{2}{c|}{\textbf{Speed-up} }
    & \multicolumn{2}{c|}{\makecell{\textbf{Avg. Inf.} \\ \textbf{Time (ms)}}} 
    & \multicolumn{2}{c|}{\makecell{\textbf{Early-Exit / Final-Exit} \\ \textbf{Avg. Inf. Time (ms)}}} 
    & \multicolumn{2}{c|}{\makecell{\textbf{CPU Avg.} \\ \textbf{(\%)}}}  
    & \multicolumn{2}{c|}{\makecell{\textbf{GPU Avg.} \\ \textbf{(\%)}}}  
    & \multicolumn{2}{c|}{\makecell{\textbf{RAM Avg.} \\ \textbf{(MB)}}} \\ \cline{3-16} 
& & \rule{0pt}{1.0\normalbaselineskip}CPU & CUDA & CPU & CUDA & CPU & CUDA & CPU & CUDA & CPU & CUDA & CPU & CUDA & CPU & CUDA \\
\hline
\multirow{9}{*}{\makecell{Jetson \\Orin\\Nano}} 
    & BASE & 1327.4 & 257.4 & 1.00x & 1.00x & 256.0 & 39.9 & - & - & 98.3 & 12.0 & - & 74.1 & 510.9 & 764.9  \\
    & PRUNE & 647.0 & 188.2 & 2.05x & 1.37x & 119.8 & 28.8 & - & - & 97.2 & 16.2 & - & 62.3 & 356.1 & 793.0\\
    & PTQ & 627.4 & \sout{742.4} & 2.12x & \sout{0.35x} & 116.1 & \sout{138.9} & - & - & 97.2 & \sout{91.8} & - & \sout{21.3} & \textcolor{good}{313.9} & \sout{548.8}\\
    & DQ & 736.2 & \sout{1015.4} & 1.80x & \sout{0.25x} & 137.9 & \sout{193.3} & - & - & 97.4 & \sout{93.2} & - & \sout{26.3} & 321.1 & \sout{570.1}\\
    & EE3 (\textit{acc-opt}) & \textcolor{bad}{1332.1} & \textcolor{bad}{238.1} & \textcolor{bad}{0.99x} & \textcolor{bad}{1.08x} & \textcolor{bad}{257.2} & \textcolor{bad}{38.3} & 73.3/268.2 & 13.1/39.8 & 98.2 & 13.0 & - & 75.1 & \textcolor{bad}{584.2} & 783.1\\
    & EE3 (\textit{inf-opt}) & \textcolor{good}{615.8} & \textcolor{good}{152.3} & \textcolor{good}{2.16x} & \textcolor{good}{1.69x} & \textcolor{good}{113.9} & \textcolor{good}{19.9} & 73.4/269.3 & 14.0/42.3 & 97.1 & 12.3 & - & 62.7 & 574.3 & 782.7\\
    & EE8 (\textit{acc-opt}) & 1311.6 & 242.4 & 1.01x & 1.06x & 253.0 & 38.0 & 122.3/269.7 & 19.6/40.4 & 98.1 & 12.3 & - & 73.1 & 547.6 & \textcolor{good}{769.8} \\
    & EE8 (\textit{inf-opt}) & 744.4 & 165.7 & 1.78x & 1.55x & 139.2 & 23.2 & 122.2/267.7 & 20.7/41.7 & 97.2 & 12.7 & - & 64.1 & 539.2 & 772.4 \\
    & EE14 (\textit{acc-opt}) & 1272.8 & 229.6 & 1.04x& 1.12x& 245.0 & 36.5 & 180.5/267.3 & 26.9/39.8 & 98.2 & 12.9 & - & 74.9 & 516.8 & 812.0 \\
    & EE14 (\textit{inf-opt}) & 953.6 & 184.4 & 1.39x & 1.40x & 181.5 & 26.7 & 181.1/267.7 & 26.7/39.7 & 97.7 & 13.1  & - & 70.1 & 510.9 & \textcolor{bad}{817.9} \\
\hline
\multirow{9}{*}{\makecell{Jetson \\ AGX\\Orin}} 
    & BASE & 635.6 & 136.6 & 1.00x & 1.00x & 120.5 & 20.9 & - & - & 97.9 & 7.9 & - & 62.6 & 440.3 & 737.4  \\
    & PRUNE & 339.6 & 131.7 & 1.87x & 1.04x & 61.1 & 19.8 & - & - & 96.3 & 7.9 & - & 63.5 & 341.8 & 752.8\\
    & PTQ &  386.6 & \sout{470.9} & 1.64x &  \sout{0.29x} & 70.6 & \sout{87.2} & - & - & 96.1 & \sout{91.3} & - & \sout{22.5} & 307.5 & \sout{488.8}\\
    & DQ & 451.6 & \sout{682.9} & 1.41x & \sout{0.20x} & 84.0 & \sout{129.5} & - & - & 96.8 & \sout{93.8} & - & \sout{27.8} & \textcolor{good}{291.5} & \sout{554.0}\\
    & EE3 (\textit{acc-opt}) & \textcolor{bad}{649.1} & \textcolor{bad}{141.8} & \textcolor{bad}{0.98x} & \textcolor{bad}{0.96x} & \textcolor{bad}{123.1} & \textcolor{bad}{22.0} & 35.4/128.4 & 7.3/22.9 & 97.8  & 7.9 & - & 67.8 & \textcolor{bad}{565.7} & \textcolor{bad}{887.2}\\
    & EE3 (\textit{inf-opt}) & \textcolor{good}{306.5} & \textcolor{good}{99.8} & \textcolor{good}{2.07x} & \textcolor{good}{1.37x} & \textcolor{good}{54.7} & \textcolor{good}{13.6} & 35.5/128.4 & 9.7/28.3 & 95.9 & 6.6 & - & 66.2 & 562.6 & 837.1\\
    & EE8 (\textit{acc-opt}) & \textcolor{bad}{649.1} & 138.0 & \textcolor{bad}{0.98x} & 0.99x & \textcolor{bad}{123.1} & 21.3 & 58.1/127.5 & 10.8/22.6 & 97.8 & 7.9 & - & 66.7 & \textcolor{bad}{565.7} & 798.7 \\
    & EE8 (\textit{inf-opt}) & 365.3 & 106.0 & 1.74x & 1.29x & 66.5 & 14.9 & 58.2/128.6 & 13.3/26.6 & 96.4 & 7.0 & - & 66.5 & 528.6 & 799.8 \\
    & EE14 (\textit{acc-opt}) & 618.7 & 133.7 & 1.03x & 1.02x & 117.1 & 20.4 & 85.8/128.0 & 14.9/22.3 & 97.7 & 7.9 & - & 64.7 & 514.9 & \textcolor{good}{770.3} \\
    & EE14 (\textit{inf-opt}) & 464.7 & 116.6 & 1.37x & 1.17x & 86.4 & 16.8 & 85.7/128.0 & 16.8/25.6 & 95.8 & 7.5 & - & 68.5 & 487.7 & 844.1 \\
\hline
\multirow{9}{*}{RPi5} 
    & BASE & 1936.4 & - & 1.00x & -& 379.9 & - & - & - & 99.3 & - & - & - & 419.7 &  - \\
    & PRUNE & \textcolor{good}{764.1} & - & \textcolor{good}{2.53x} & -& \textcolor{good}{145.2} & - & - & - & 98.6 & - & - & - & 337.2 & -\\
    & PTQ & 1031.6 & - & 1.88x &- & 198.6 & - & - & - & 98.7 & - & - & - & 301.2 & -\\
    & DQ & 1149.2 & - & 1.68x &- & 222.1 & - & - & - & 98.9 & - & - & - & \textcolor{good}{290.2} & -\\
    & EE3 (\textit{acc-opt}) & 1847.4 & - & 1.05x & -& 362.2 & - & 112.4/377.2 & - & 99.3 & - & - & - & \textcolor{bad}{557.2} & -\\
    & EE3 (\textit{inf-opt}) & 862.9 & - & 2.24x &- & 165.1 & - & 111.8/369.4 & - & 98.7 & - & - & - & 554.9 & -\\
    & EE8 (\textit{acc-opt}) & \textcolor{bad}{1872.2} & - & \textcolor{bad}{1.03x} &- & \textcolor{bad}{367.0} & - & 181.5/390.7 & - & 99.2 & - & - & - & 522.1 & - \\
    & EE8 (\textit{inf-opt}) & 1064.3 & - & 1.82x& -& 205.3 & - & 181.2/386.8 & - & 98.9 & - & - & - & 520.9 & - \\
    & EE14 (\textit{acc-opt}) & 1833.6 & - & 1.06x &- & 359.2 & - & 268.8/390.3 & - & 99.2 & - & - & - & 454.7 & - \\
    & EE14 (\textit{inf-opt}) & 1384.0 & - & 1.40x & -& 269.2 & - & 268.8/387.9 & - & 99.0 & - & - & - & 477.6 & - \\
\hline
\end{tabular}
\end{center}
\end{table*}

\begin{table*}[t]
\caption{Edge device performance metrics utilization EfficientNet-B2}
\label{tab:performance_efficientnet}
\begin{center}
\tiny
\begin{tabular}{|c|c|c@{\hspace{1mm}}c|c@{\hspace{1mm}}c|c@{\hspace{1mm}}c|c@{\hspace{1mm}}c|c@{\hspace{1mm}}c|c@{\hspace{1mm}}c|c@{\hspace{1mm}}c|}
\hline
\multirow{2}{*}{\textbf{Device}} & \multirow{2}{*}{\textbf{Model}}
    & \multicolumn{2}{c|}{\makecell{\textbf{Total Inf.} \\ \textbf{Time (s)}}}  
    & \multicolumn{2}{c|}{\textbf{Speed-up} }
    & \multicolumn{2}{c|}{\makecell{\textbf{Avg. Inf.} \\ \textbf{Time (ms)}}} 
    & \multicolumn{2}{c|}{\makecell{\textbf{Early-Exit / Final-Exit} \\ \textbf{Avg. Inf. Time (ms)}}} 
    & \multicolumn{2}{c|}{\makecell{\textbf{CPU Avg.} \\ \textbf{(\%)}}}  
    & \multicolumn{2}{c|}{\makecell{\textbf{GPU Avg.} \\ \textbf{(\%)}}}  
    & \multicolumn{2}{c|}{\makecell{\textbf{RAM Avg.} \\ \textbf{(MB)}}} \\ \cline{3-16} 
& & \rule{0pt}{1.0\normalbaselineskip}CPU & CUDA & CPU & CUDA & CPU & CUDA & CPU & CUDA & CPU & CUDA & CPU & CUDA & CPU & CUDA \\
\hline
\multirow{9}{*}{\makecell{Jetson \\Orin\\Nano}} 
    & BASE  & \textbf{402.4} & \textbf{140.8} & \textbf{1.00x} & \textbf{1.00x} & \textbf{71.1} & \textbf{19.1}& - & - & \textbf{96.6} & \textbf{14.5} & - & \textbf{65.0} & \textbf{249.9} & \textbf{765.3}\\
    & PRUNE  & 362.3 & 141.4 & 1.11x & 0.99x & 63.1 & 19.4 & - & - & 96.3 & 14.7 & - & 66.6 & 234.1 & 761.5\\
    & PTQ  & 273.3 & \sout{291.6} & 1.47x & \sout{0.48x} & 45.3 & \sout{48.9} & - & - & 94.8 & \sout{89.2} & - & \sout{36.8} & 245.6 & \sout{541.8}\\
    & DQ  & \textcolor{bad}{522.9} & \sout{637.3} & \textcolor{bad}{0.77x} & \sout{0.22x} & \textcolor{bad}{95.3} & \sout{118.0} & - & - & 97.1 & \sout{93.0} & - & \sout{24.4} & \textcolor{good}{219.4} & \sout{529.1}\\ 
    & EE3 (\textit{acc-opt}) & 446.1 & 152.1 & 0.90x & 0.93x & 79.9 & 21.6 & 13.6/79.9 & 6.1/21.6 & 96.8 & 14.2 & - & 62.5 & \textcolor{bad}{285.6} & 781.3\\
    & EE3 (\textit{inf-opt}) & \textcolor{good}{246.3} & \textcolor{good}{120.5} & \textcolor{good}{1.63x} & \textcolor{good}{0.57x} & \textcolor{good}{40.0} & \textcolor{good}{15.1} & 14.2/79.4 & 7.5/26.9 & 95.0 & 12.7 & - & 60.4 & 285.4 & \textcolor{bad}{811.0}\\
    & EE8 (\textit{acc-opt}) & 423.1 & 156.8 & 0.95x &  0.90x & 75.3 & 22.5 & 32.3/78.8 & 12.6/23.3 & 96.8 & 13.9 & - & 63.8 & 285.3 & \textcolor{good}{778.5}\\
    & EE8 (\textit{inf-opt}) & 264.0 & 132.4 & 1.52x & 1.06x & 43.5 & 17.1 & 32.3/78.7 & 14.1/26.5 & 95.1 & 11.9 & - & 61.0 & 279.7 & 780.9\\
    & EE15 (\textit{acc-opt}) & 368.7 & 151.5 & 1.09x & 0.93x & 64.4 & 21.3 & 49.6/79.3 & 18.2/24.5 & 96.3 & 12.9 & - & 63.2 & 276.1 & 795.0\\
    & EE15 (\textit{inf-opt}) & 294.6 & \textcolor{bad}{188.1} & 1.37x & \textcolor{bad}{0.75x} & 49.7 & \textcolor{bad}{23.8} & 49.2/78.9 & 23.7/29.8 & 95.4 & 10.9 & - & 63.0 & 279.5 & 806.9\\
\hline
\multirow{9}{*}{\makecell{Jetson \\ AGX\\Orin}} 
    & \textbf{BASE} & \textbf{276.4} & \textbf{104.0} & \textbf{1.00x} & \textbf{1.00x} & \textbf{48.5} & \textbf{14.3} & - & - & \textbf{96.1} & \textbf{7.3} & - & \textbf{66.1} & \textbf{236.4} & \textbf{781.0}\\
    & PRUNE & 245.2 & 101.8 & 1.13x & 1.02x & 42.3 & 14.0 & - & - & 95.6 & 7.5 & - & 65.9 & 233.3 & \textcolor{good}{779.8}\\
    & PTQ & 188.1 & \sout{221.8} & 1.47x & \sout{0.63x} & 31.0 & \sout{37.4} & - & - & 93.7 & \sout{86.1} & - & \sout{36.4} & 229.8 & \sout{534.4}\\  
    & DQ & \textcolor{bad}{357.3} & \sout{472.6} & \textcolor{bad}{0.77x} & \sout{0.22x} & \textcolor{bad}{64.8} & \sout{87.7} & - & - & 96.6 & \sout{92.9} & - & \sout{23.8} & \textcolor{good}{210.6} & \sout{492.4}\\
    & EE3 (\textit{acc-opt}) & 302.8 & \textcolor{bad}{111.9} & 0.91x & \textcolor{bad}{0.93x} & 53.8 & \textcolor{bad}{16.1} & 9.6/53.8 & 4.4/16.1 & 96.1 & 7.4 & - & 66.4 & \textcolor{bad}{270.6} & 799.4\\
    & EE3 (\textit{inf-opt}) & \textcolor{good}{169.7} & \textcolor{good}{92.6} & \textcolor{good}{1.63x} & \textcolor{good}{1.12x} & \textcolor{good}{27.3} & \textcolor{good}{11.9} & 9.7/54.0 & 5.9/21.2 & 93.7 & 6.3 & - & 58.9 & 267.5 & 793.6\\
    & EE8 (\textit{acc-opt}) & 288.0 & 111.7 & 0.96x & 0.93x & 50.8 & 16.1 & 22.3/53.1 & 8.3/16.7 & 96.0 & 7.1 & - & 61.7 & 258.5 & 792.9\\
    & EE8 (\textit{inf-opt}) & 181.6 & 100.4 & 1.52x & 1.04x & 29.7 & 13.1 & 22.2/53.1 & 10.4/21.5 & 94.1 & 6.1 & - & 62.1 & 256.8 & 790.5\\
    & EE15 (\textit{acc-opt}) & 255.3 & 103.5 & 1.08x & 1.00x & 44.4 & 14.4 & 34.4/54.4 & 12.0/16.9 & 95.8 & 6.8 & - & 65.3 & 262.2 & \textcolor{bad}{800.8}\\
    & EE15 (\textit{inf-opt}) & 202.7 & 104.1 & 1.36x & 0.99x & 34.0 & 14.3 & 33.7/53.3 & 14.3/20.6 & 94.7 & 6.2 & - & 62.7 & 268.8 & 792.6\\
\hline
\multirow{9}{*}{\makecell{RPi5}} 
    & BASE & \textbf{525.1} & - & \textbf{1.00x} & - & \textbf{97.3} & - & - & - & \textbf{98.1} & - & - & - & \textbf{226.1} & -\\
    & PRUNE & 452.4 & - & 1.16x & - & 82.9 & - & - & - & 98.0 & - & - & - & \textcolor{good}{221.5} & -\\
    & PTQ & 459.7 & - & 0.98x & - &  84.4 & - & - & - & 97.8 & - & - & - & 224.0 & - \\
    & DQ & \textcolor{bad}{653.7} & - & \textcolor{bad}{0.80x} & - & \textcolor{bad}{123.0} & - & - & - & 98.3 & - & - & - & 208.3 & -\\
    & EE3 (\textit{acc-opt}) & 576.2 & - & 0.91x & - & 107.6 & - & 30.9/107.7 & - & 98.3 & - & - & - & \textcolor{bad}{254.6} & - \\
    & EE3 (\textit{inf-opt}) & \textcolor{good}{345.8} & - & \textcolor{good}{1.52x} & - & \textcolor{good}{61.4} & - & 31.1/107.6 & - & 97.2 & - & - & - & 254.4 & -\\
    & EE8 (\textit{acc-opt}) & 537.5 & - & 0.98x & - & 99.9 & - & 57.2/103.4 & - & 98.2 & - & - & - & 248.7 & -\\
    & EE8 (\textit{inf-opt}) & 477.4 & - & 1.09x & - & 87.7 & - & 57.2/104.2 & - & 97.9 & - & - & - & 249.5 & -\\
    & EE15 (\textit{acc-opt}) & 480.7 & - & 1.09x & - & 88.4 & - & 74.1/102.8 & - & 97.9 & - & - &  - & 252.4 & -\\
    & EE15 (\textit{inf-opt}) & 410.4 & - & 1.27x & - & 74.2 & - & 73.8/103.3& - & 97.5 & - & - & - & 252.6 & -\\
\hline
\end{tabular}
\end{center}
\end{table*}

\begin{table*}[ht!]
\caption{Edge device performance metrics utilization MobileNet-V2}
\label{tab:performance_mobilenet}
\begin{center}
\tiny
\begin{tabular}{|c|c|c@{\hspace{1mm}}c|c@{\hspace{1mm}}c|c@{\hspace{1mm}}c|c@{\hspace{1mm}}c|c@{\hspace{1mm}}c|c@{\hspace{1mm}}c|c@{\hspace{1mm}}c|}
\hline
\multirow{2}{*}{\textbf{Device}} & \multirow{2}{*}{\textbf{Model}}
    & \multicolumn{2}{c|}{\makecell{\textbf{Total Inf.} \\ \textbf{Time (s)}}}  
    & \multicolumn{2}{c|}{\textbf{Speed-up} }
    & \multicolumn{2}{c|}{\makecell{\textbf{Avg. Inf.} \\ \textbf{Time (ms)}}} 
    & \multicolumn{2}{c|}{\makecell{\textbf{Early-Exit / Final-Exit} \\ \textbf{Avg. Inf. Time (ms)}}} 
    & \multicolumn{2}{c|}{\makecell{\textbf{CPU Avg.} \\ \textbf{(\%)}}}  
    & \multicolumn{2}{c|}{\makecell{\textbf{GPU Avg.} \\ \textbf{(\%)}}}  
    & \multicolumn{2}{c|}{\makecell{\textbf{RAM Avg.} \\ \textbf{(MB)}}} \\ \cline{3-16} 
& & \rule{0pt}{1.0\normalbaselineskip}CPU & CUDA & CPU & CUDA & CPU & CUDA & CPU & CUDA & CPU & CUDA & CPU & CUDA & CPU & CUDA \\
\hline
\multirow{9}{*}{\makecell{Jetson \\Orin\\Nano}} 
    & \textbf{BASE} & \textbf{179.1} & \textbf{116.7} & \textbf{1.00x} & \textbf{1.00x} &  \textbf{26.6} & \textbf{13.7} & - & - & \textbf{94.3} & \textbf{12.4} & - & \textbf{54.1} & \textbf{209.8} & \textbf{739.9}\\
    & PRUNE & 155.9 & 106.1 & 1.15x & 1.10x & 22.0 & 12.0 & - & - & 93.2 & 12.8 & - & 51.1 & 202.9 & \textcolor{good}{685.1}\\
    & PTQ & \textcolor{good}{97.0} & \sout{111.9} & \textcolor{good}{1.85x} & \sout{1.04x} & \textcolor{good}{10.2} & \sout{13.2} & - & - & 90.0 & \sout{82.5} & - & \sout{6.7} & \textcolor{good}{202.5} & \sout{372.0} \\
    & DQ & \textcolor{bad}{234.7} & \sout{305.8} & \textcolor{bad}{0.76x} & \sout{0.38x} & \textcolor{bad}{37.7} & \sout{51.6} & - & - & 95.0 & \sout{91.6} & - & \sout{23.8} & 212.5 & \sout{509.6} \\ 
    & EE3 (\textit{acc-opt}) & 211.0 & \textcolor{bad}{130.3} & 0.85x & \textcolor{bad}{0.89x} &  32.9 & \textcolor{bad}{16.1} & 8.5/33.1 & 6.1/16.2  & 94.5 & 12.5 & - & 56.6 & 213.1 & \textcolor{bad}{764.9} \\
    & EE3 (\textit{inf-opt}) & 145.0 & \textcolor{good}{101.3} & 1.24x & \textcolor{good}{1.15x} & 19.8 & \textcolor{good}{10.7} & 8.5/33.5 & 6.5/16.4 & 92.7 & 12.5 & - & 49.1 & \textcolor{bad}{214.2} & 751.0\\
    & EE8 (\textit{acc-opt}) & 202.1 & 117.7 & 0.89x & 0.98x & 31.1 & 14.1 &17.2/32.4 & 8.8/14.6& 96.2 & 12.7 & - & 57.0 & 201.4 & 746.9 \\
    & EE8 (\textit{inf-opt}) & 147.1 & 103.4 & 1.22x & 1.23x& 20.3 & 11.1 & 16.3/31.5 & 9.5/15.4 & 93.2 & 12.4 & - & 47.4 & 208.9 & 744.9 \\
    & EE13 (\textit{acc-opt}) & 188.6 & 121.6 & 0.95x& 0.96x & 28.4 & 14.7 & 22.9/29.5 & 11.5/15.4 & 97.1 & 12.5 & - & 54.6 & 211.9 & 749.6\\
    & EE13 (\textit{inf-opt}) & 153.1 & 104.4 & 1.17x& 1.12x & 21.4 & 11.5 & 21.3/27.9 & 11.5/15.1 & 92.7 & 12.6 & - & 53.3 & 210.9 & 685.9  \\
\hline
\multirow{9}{*}{\makecell{Jetson \\ AGX\\Orin}} 
    & \textbf{BASE} & \textbf{124.1} & \textbf{74.5} & \textbf{1.00x} & \textbf{1.00x} & \textbf{18.2} & \textbf{8.4} & - & - & \textbf{92.9} & \textbf{6.5} & - & \textbf{57.1} & \textbf{189.8} & \textbf{711.9}\\
    & PRUNE& 108.1 & 69.7 & 1.14x & 1.07x & 15.1 & 7.7 & - & - & 91.9 & 6.7 & - & 54.5 & \textcolor{bad}{196.5} & 774.3\\
    & PTQ & \textcolor{good}{98.3} & \sout{124.0} & \textcolor{good}{1.27x} & \sout{0.60x} & \textcolor{good}{13.0} & \sout{17.9} & - & - & 89.4 & \sout{84.3} & - & \sout{25.3} & \textcolor{good}{193.3} & \sout{449.9} \\  
    & DQ & \textcolor{bad}{169.5} & \sout{235.1} & \textcolor{bad}{0.73x} & \sout{0.32x} & \textcolor{bad}{27.1} & \sout{40.0} & - & - & 93.6 & \sout{91.3} & - & \sout{25.2} & 200.5 & \sout{461.8} \\ 
    & EE3 (\textit{acc-opt}) & 139.0 & \textcolor{bad}{77.9} & 0.89x & \textcolor{bad}{0.96x} & 21.1 & \textcolor{bad}{9.3}  & 7.1/21.2 & 3.5/9.3 & 92.8 & 6.6 & - & 53.4 & 203.4 & 783.0 \\
    & EE3 (\textit{inf-opt}) & 98.6 & \textcolor{good}{62.8} & 1.26x & \textcolor{good}{1.19x} & 13.2 & \textcolor{good}{6.3} & 6.6/21.2 & 3.5/9.6 & 92.2 & 6.6 & - & 40.8 & 206.2   & 712.4\\
    & EE8 (\textit{acc-opt}) & 125.9 & \textcolor{bad}{77.9} & 0.99x & \textcolor{bad}{0.96x} & 18.6 & 9.2 & 11.7/19.2 & 5.4/9.5 & 97.3 & 6.4 & - & 55.3 &  202.9  & \textcolor{good}{711.8}\\
    & EE8 (\textit{inf-opt}) & 100.0 & 63.4 & 1.24x& 1.18x & 13.5 & 6.4 & 11.6/19.0 & 5.4/9.2 & 92.7 & 6.5 & - & 46.9 & 200.6 & \textcolor{bad}{788.8} \\
    & EE13 (\textit{acc-opt}) & 126.0 & 76.1 & 0.98x& 0.98x & 18.6 & 9.0 & 15.3/19.2 & 7.2/9.3 & 96.9 & 6.6 & - & 53.7 & 204.3 & 777.5 \\
    & EE13 (\textit{inf-opt}) & 108.3 & 66.9 & 1.15x & 1.11x &  15.2 & 7.1 & 15.2/19.2 & 7.0/9.7 & 91.6 & 6.5 & - & 46.4 & 210.9 & 775.0  \\
\hline
\multirow{9}{*}{\makecell{RPi5}} 
    & \textbf{BASE} & \textbf{209.2} & - & \textbf{1.00x}&- & \textbf{34.3} & - & - & - & \textbf{96.1} & - & - & - & \textbf{191.2} & -\\
    & PRUNE & 168.7 & - & 1.24x &- & 26.5 & - & - & - & 95.9 & - & - & - & 183.3 & -\\
    & PTQ & 192.5 & - & 1.09x & -& 31.1 & - & - & - & 95.7 & - & - & - & 183.8 & -\\
    & DQ & \textcolor{bad}{299.1} & - & \textcolor{bad}{0.70x} & -& \textcolor{bad}{52.3} & - & - & - & 97.2 & - & - & - & 194.5 & -\\
    & EE3 (\textit{acc-opt}) & 227.7 & - & 0.92x & -&  38.1 & - & 15.4/38.2 & - & 96.6 & - & - & - &  \textcolor{bad}{195.2}  & -\\
    & EE3 (\textit{inf-opt}) & \textcolor{good}{167.0} & - & \textcolor{good}{1.25x} & - & \textcolor{good}{25.8} & - & 15.5/38.2 & - & 95.0 & - & - & - &  194.4  & -\\
    & EE8 (\textit{acc-opt}) & 224.7 & - & 0.93x &- & 37.6 & - & 24.8/38.8 & - & 96.6 & - & - & - & 192.8 & -\\
    & EE8 (\textit{inf-opt}) & 179.8 & - & 1.16x& - & 28.4 & - & 24.7/38.7 & - & 95.3 & - & - & - & \textcolor{good}{192.7}  & -\\
    & EE13 (\textit{acc-opt}) & 224.0 & - & 0.93x & - & 37.5 & - & 29.5/39.1 & - & 96.4 & - & - & - &  194.0  & -\\
    & EE13 (\textit{inf-opt}) & 185.2 & - &1.13x  & - & 29.6 & - & 29.5/39.4 & - & 95.5 & - & - & - &  194.6  & -\\
\hline
\end{tabular}
\end{center}
\end{table*}

\begin{table*}[ht!]
\caption{Edge device performance metrics utilization ShuffleNet-V2}
\label{tab:performance_shufflenet}
\begin{center}
\tiny
\begin{tabular}{|c|c|c@{\hspace{1mm}}c|c@{\hspace{1mm}}c|c@{\hspace{1mm}}c|c@{\hspace{1mm}}c|c@{\hspace{1mm}}c|c@{\hspace{1mm}}c|c@{\hspace{1mm}}c|}
\hline
\multirow{2}{*}{\textbf{Device}} & \multirow{2}{*}{\textbf{Model}}
    & \multicolumn{2}{c|}{\makecell{\textbf{Total Inf.} \\ \textbf{Time (s)}}}  
    & \multicolumn{2}{c|}{\textbf{Speed-up} }
    & \multicolumn{2}{c|}{\makecell{\textbf{Avg. Inf.} \\ \textbf{Time (ms)}}} 
    & \multicolumn{2}{c|}{\makecell{\textbf{Early-Exit / Final-Exit} \\ \textbf{Avg. Inf. Time (ms)}}} 
    & \multicolumn{2}{c|}{\makecell{\textbf{CPU Avg.} \\ \textbf{(\%)}}}  
    & \multicolumn{2}{c|}{\makecell{\textbf{GPU Avg.} \\ \textbf{(\%)}}}  
    & \multicolumn{2}{c|}{\makecell{\textbf{RAM Avg.} \\ \textbf{(MB)}}} \\ \cline{3-16} 
& & \rule{0pt}{1.0\normalbaselineskip}CPU & CUDA & CPU & CUDA & CPU & CUDA & CPU & CUDA & CPU & CUDA & CPU & CUDA & CPU & CUDA \\
\hline
\multirow{9}{*}{\makecell{Jetson \\Orin\\Nano}} 
    & BASE  & 99.6 & 85.4 & 1.00x & 1.00x  & 10.9 & 8.4 & - & - & 90.8 & 16.5 & - & 41.4 & 193.6 & 666.1\\
    &PRUNE  & - & - & & & - & - & - & - & - & - & - & - & - & -\\
    & PTQ  & \textcolor{bad}{152.0} & \sout{116.8} & \textcolor{bad}{0.66x} & \sout{0.73x} & \textcolor{bad}{21.4} & \sout{14.5} & - & - & 93.3 & \sout{14.9} & - & \sout{60.0} & \textcolor{good}{191.3} & \sout{669.3}\\
    & DQ  & 121.5 & \sout{160.0} & 0.82x & \sout{0.53x} & 15.3 & \sout{23.0} & - & - & 92.2 & \sout{89.9} & - & \sout{25.2} & 193.8 & \sout{447.8}\\ 
    & EE2 (\textit{acc-opt}) & 125.6 & \textcolor{bad}{98.6} & 0.79x & \textcolor{bad}{0.87x} & 16.0 & \textcolor{bad}{10.9} & 4.4/16.1 & 3.5/10.9 & 92.7 & 16.2 & - & 41.1 & 196.1 & \textcolor{good}{674.2} \\
    & EE2 (\textit{inf-opt}) & 92.3 & 79.2 & 1.08x & 1.07x & 9.5 & 7.1 & 4.1/15.0 & 3.4/10.9 & 92.0 & 16.0 & - & 32.6 & 197.3 & 677.0\\
    & EE5 (\textit{acc-opt})  & 108.5 & 89.4 & 0.92x & 0.96x & 12.6 & 9.1 & 6.4/14.2 & 4.6/10.2 & 95.2 & 16.4 & - & 39.7 & \textcolor{bad}{199.6} & \textcolor{bad}{685.5}\\
    & EE5 (\textit{inf-opt})  & \textcolor{good}{90.5} & \textcolor{good}{77.9} & \textcolor{good}{1.10x} & \textcolor{good}{1.10x} & \textcolor{good}{9.0} & \textcolor{good}{6.9} & 6.3/13.5 & 4.6/10.5 & 91.9 & 16.2 & - & 35.3 & 195.2 & 685.0  \\
    & EE8 (\textit{acc-opt})  & 102.5 & 86.4 & 0.97x & 0.99x& 11.5 & 8.5 & 7.9/13.6 & 5.7/10.2 & 94.5 & 16.4 & - & 40.5 & 196.7 & 678.9\\
    & EE8 (\textit{inf-opt}) & 91.9 & 79.2 & 1.08x & 1.08x & 9.3 & 7.1 & 7.6/13.2 & 5.7/10.4 & 91.9 & 16.5 & - & 41.6 & 191.7 & 679.1  \\
\hline
\multirow{9}{*}{\makecell{Jetson \\ AGX\\Orin}} 
    & BASE & 68.5 & 64.6 & 1.00x & 1.00x& 7.3 & 6.6 & - & - & 88.3 & 7.7 & - & 47.9 & 181.3 & 684.3\\
    & PRUNE & - & - & -& -& - & - & - & - & - & - & - & - & - & -\\
    & PTQ & \textcolor{good}{58.3} & \sout{76.2} & \textcolor{good}{1.17x} & \sout{0.85x} &  \textcolor{good}{5.2} & \sout{8.7} & - & - & 86.5 & \sout{78.1} & - & \sout{19.7} & \textcolor{good}{183.8} & \sout{370.3} \\  
    & DQ & 85.9 & \sout{127.3} & 0.70x & \sout{0.51x}& 10.8 & \sout{18.8} & - & - & 90.7 & \sout{88.4} & - & \sout{26.3} & 178.8 & \sout{444.0} \\
    & EE2 (\textit{acc-opt}) & \textcolor{bad}{87.4} & \textcolor{bad}{71.3} & \textcolor{bad}{0.78x} & \textcolor{bad}{0.91x} & \textcolor{bad}{11.0} & \textcolor{bad}{8.0} & 2.7/11.0 & 2.7/8.0 & 90.4 & 7.9 & - & 46.9 & 186.4 & 702.3 \\
    & EE2 (\textit{inf-opt}) & 63.4 & 56.5 & 1.08x & 1.14x & 6.3 & \textcolor{good}{5.1} & 2.7/10.1 & 2.4/7.9 & 89.8 & 7.9 & - & 37.4 & 193.1 & 694.2\\
    & EE5 (\textit{acc-opt}) & 76.2 & 65.5 & 0.90x & 0.99x & 8.8 & 6.9 & 4.4/9.9 & 3.6/7.7 & 91.9 & 8.0 & - & 41.9 & \textcolor{bad}{192.9} & \textcolor{good}{689.8} \\
    & EE5 (\textit{inf-opt}) & 63.8 & 56.7 & 1.07x & 1.14x & 6.1 & \textcolor{good}{5.1} & 4.3/9.1 & 3.5/7.7 & 89.9 & 7.8 & - & 41.9 & 187.0 & \textcolor{good}{704.3}  \\
    & EE8 (\textit{acc-opt}) & 71.0 & 62.9 & 0.96x & 1.03x & 7.8 & 6.4 & 5.3/9.2 & 4.4/7.6 & 92.5 & 7.8 & - & 46.8 & 184.0 & 696.3 \\
    & EE8 (\textit{inf-opt}) & 62.9 & \textcolor{good}{56.2} & 1.09x & \textcolor{good}{1.15x} & 6.2 & \textcolor{good}{5.1} & 5.3/9.0 & 4.2/7.6 & 88.7 & 8.0 & - & 39.1 & 187.8 & 698.2  \\
\hline
\multirow{9}{*}{RPi5}
    & BASE & 87.3 & - & 1.00x & - &  10.2 & - & - & - & 92.2 & - & - & - &  179.4 & -\\
    & PRUNE & - & - & - &-  &  - & - & - & - & - & - & - & - & - & -\\
    & PTQ & \textcolor{good}{73.3} & - & \textcolor{good}{1.19x} & -& \textcolor{good}{7.1} & - & - & - & 89.7 & - & - & - & 177.1 & -\\
    & DQ & 103.9 & - & 0.84x & -& 13.1 & - & - & - & 92.5 & - & - & - & 177.9 & -\\
    & EE2 (\textit{acc-opt}) & \textcolor{bad}{114.6} & - & \textcolor{bad}{0.76x} & - & \textcolor{bad}{15.5} & - & 4.7/15.5 & - & 93.8 & - & - & - & 182.9 & -\\
    & EE2 (\textit{inf-opt}) & 86.5 & - & 1.01x & - & 9.7 & - & 4.4/15.2 & - & 90.4 & - & - & - & \textcolor{bad}{182.8} & -\\
    & EE5 (\textit{acc-opt}) & 103.0 & - & 0.85x & -& 13.1 & - & 6.4/14.9 & - & 92.5 & - & - & - & 183.1 & -\\
    & EE5 (\textit{inf-opt}) & 86.1 & - & 1.01x & -& 9.6 & - & 6.4/14.8 & - & 90.5 & - & - & - & 182.1 & -\\
    & EE8 (\textit{acc-opt}) & 97.1 & - & 0.90x & -& 12.0 & - & 7.7/14.6 & - & 92.2 & - & - & - &  \textcolor{good}{172.1}  & -\\
    & EE8 (\textit{inf-opt}) & 84.1 & - & 1.04x & -& 9.4 & - & 7.6/14.5 & - & 90.9 & - & - & - & 173.0 & -\\
\hline
\end{tabular}
\end{center}
\end{table*}

Based on the analysis of model size and accuracy, this section investigates how the proposed optimization techniques translate into practical inference efficiency when implemented on resource-constrained edge devices. Specifically, we evaluate inference latency and runtime resource utilization on three representative platforms (see section~\ref{sec:evalu_metrics}).

For each device and architecture, we report total and per-sample inference latency, along with average CPU, GPU, and RAM usage, enabling a comprehensive assessment of computational efficiency. For early-exit models, we further decompose the inference behavior by distinguishing samples that exit at intermediate stages from those that reach the final classifier. This allows direct quantification of the latency savings achieved and highlights the dynamic nature of early-exit inference compared to static optimization methods.

The results provide insight into how architectural features and optimization strategies interact under real deployment conditions. A comprehensive summary is presented in Tables~\ref{tab:performance_resnet}–\ref{tab:performance_shufflenet}.

\textbf{Pruning} has a clear, backend-dependent impact on inference speed across all evaluated architectures. On CPU-based edge devices, pruning can achieve substantial acceleration for large models such as ResNet-152, yielding up to 2.53x acceleration on the Raspberry Pi 5. Compact architectures, including MobileNet-V2 and EfficientNet-B2, also benefit from pruning on the CPU, although the observed gains are more modest, typically 1.1x to 1.25x.

On GPU-based platforms, pruning produces more moderate acceleration across all architectures. Large models, such as ResNet-152, achieve accelerations of up to 1.37x on Orin Nano, while compact architectures experience smaller but still measurable gains, typically not exceeding 1.15x. On high-end platforms, such as AGX Orin, the relative acceleration is further reduced. This limited acceleration on the GPU persists even when applying structured pruning with hardware-aligned channel granularities (e.g., multiples of 32), as standard GPU runtimes are optimized for dense operations and cannot fully leverage the modifications introduced by pruning \cite{DBLP:journals/corr/abs-1802-10280}.

In addition, the reduction in acceleration observed in compact models is also related to the lower pruning rates applied to these architectures. As discussed in the Appendix~\ref{app:pruning}, aggressive pruning in compact models leads to rapid accuracy degradation, limiting the level of pruning that can be achieved. Consequently, the limited degree of pruning directly restricts the inference acceleration that can be achieved, even when pruning remains effective in absolute terms.

In terms of computational resource utilization, structured pruning reveals a subtle but consistent pattern across architectures. In CPU-only executions, pruning typically correlates with a slight reduction in CPU utilization (0.3–1.1\%, e.g., ResNet-152: 98.3\% to 97.2\%). Conversely, when leveraging CUDA, CPU usage shows a marginal increase (0–0.4\%). While these variations are slight and near typical measurement noise, their consistent direction suggests that pruning alters the computational balance: as heavy computation is offloaded to the GPU, the CPU's relative workload may shift toward data management and kernel coordination. This indicates that pruning's primary benefits are reduced memory footprint and faster inference, rather than substantial reductions in instantaneous processor load.

Regarding GPU utilization under CUDA execution, pruning produces mixed effects that vary by architecture and device. MobileNet-V2 shows consistent reductions (2.6–3.0\% lower GPU usage on both Jetson devices), while EfficientNet-B2 exhibits minimal fluctuations (±1.6\%). ResNet-152 presents divergent behavior: an 11.8\% decrease on Jetson Orin Nano contrasted with a 0.9\% increase on AGX Orin. These inconsistent patterns suggest that pruning's impact on GPU load depends on complex interactions between model structure, sparsity patterns, and device-specific scheduling optimizations, making it less predictable than pruning's consistent benefits for model size and inference latency.

When it comes to memory footprint, pruning consistently reduces RAM consumption in CPU-only execution across all architectures and devices. The savings are most pronounced for large models, with ResNet-152 showing substantial reductions (e.g., 510.9 MB to 356.1 MB on Orin Nano, a 30\% decrease). For compact architectures like EfficientNet-B2 and MobileNet-V2, reductions are smaller but still measurable (typically 2–5\%). Interestingly, in GPU-accelerated execution, the RAM pattern becomes inconsistent: while MobileNet-V2 shows reduced memory usage on Orin Nano (739.9 MB to 685.1 MB), ResNet-152 and EfficientNet-B2 exhibit negligible changes or even slight increases. This suggests that the GPU memory management and runtime overhead can sometimes offset or outweigh the memory savings from model sparsity, particularly when the baseline model is already memory-efficient.

In summary, pruning shows a clear hierarchy of efficiency: it is most effective for large models on CPUs, providing substantial speed increases (up to 2.53x) and consistent memory savings. In compact architectures and GPU implementations, the benefits are more limited due to lower pruning tolerance and the absence of scatter-sensitive cores in standard runtimes. However, they remain appreciable (1.1x-1.37x speed-up). Pruning reliably reduces model size and inference latency, and its principal value lies in structural efficiency rather than computational load reduction.

\textbf{Quantization} shows distinct performance trends depending on both the model architecture and the quantization strategy employed. A key observation that guides the remainder of this analysis is the severe performance regression observed when quantized models are executed on CUDA-based backends. This is because CUDA does not support many quantized layers directly. In practice, unsupported quantized layers are silently offloaded to the CPU for execution, while the remaining operations run on the GPU. This creates a hybrid CPU-GPU execution pipeline that introduces significant synchronization overhead and memory transfer bottlenecks, often resulting in higher latency than pure CPU inference \cite{Lin2024QServe:, Kim2024QUICK:, jain2020efficientexecutionquantizeddeep}. Consequently, the discussion focuses on CPU-based inference, where quantization is effectively supported. 

In this setting, \gls{ptq} quantization efficiency depends primarily on the model size and the technique used. In large architectures such as ResNet-152, both \gls{ptq} and \gls{dq} offer consistent acceleration across all devices, although \gls{ptq} outperforms \gls{dq} (e.g., 2.12x vs. 1.80x on Orin Nano). For medium-sized, efficiency-optimized models (EfficientNet-B2, MobileNet-V2), a clear divergence is observed: \gls{ptq} maintains speed (1.27x–1.85x), while \gls{dq} consistently degrades performance (0.70x–0.77x). This is explained by the fundamental difference between the two techniques: \gls{ptq} pre-calculates the activation scales during calibration, without introducing runtime overhead, while \gls{dq} must calculate them dynamically for each input. In very compact architectures such as ShuffleNet-V2, the gains are more modest and depend on the device, as \gls{ptq} typically provides slight accelerations (1.17x-1.19x on AGX Orin/RPi5), while \gls{dq} again falls behind (0.70x-0.84x). These results confirm that \gls{ptq} is more reliable and beneficial across the spectrum of architectures.

This performance divergence is also reflected in CPU utilization, though with architecture-specific variation. For ResNet-152, both \gls{ptq} and \gls{dq} reduce CPU usage (e.g., Orin Nano: 98.3\% to 97.2\% and 97.4\%), correlating with their respective speed-ups (2.12x and 1.80x). For mid-sized efficient models, \gls{ptq} consistently lowers CPU utilization (MobileNet-V2: 94.3\% to 90.0\%) while \gls{dq} increases it (95.0\%)—directly aligning with PTQ's acceleration (1.85x) and \gls{dq}'s slowdown (0.76x). ShuffleNet-V2 deviates from this pattern, exhibiting less predictable behavior consistent with its mixed quantization speed-up results.

Memory footprint reveals an architecture-dependent trade-off between \gls{ptq} and \gls{dq}. For large, parameter-dense models like ResNet-152, \gls{dq} consistently achieves lower RAM usage than \gls{ptq} (e.g., Orin Nano: 321.1 MB vs. 313.9 MB), likely because its statically quantized weights dominate memory, and \gls{dq}'s weight compression matches \gls{ptq} without needing activation calibration buffers. EfficientNet-B2 behaves as an intermediate architecture that follows the large-model pattern, with \gls{dq} consistently outperforming \gls{ptq} by a significant margin (Orin Nano: \gls{ptq} 245.6 MB vs. \gls{dq} 219.4 MB, a 10.7\% advantage). For compact architectures like MobileNet-V2 and ShuffleNet-V2, the trend reverses. On MobileNet-V2, \gls{ptq} consistently outperforms \gls{dq} in memory savings (Orin Nano: \gls{ptq} 202.5 MB vs. \gls{dq} 212.5 MB). ShuffleNet-V2 shows more minor, device-dependent differences (<2.5\%). This reversal occurs because in memory-constrained models, \gls{dq}'s runtime activation buffers and intermediate tensor storage introduce measurable overhead that outweighs its weight compression benefits. In contrast, \gls{ptq}'s precomputed scales eliminate this runtime memory footprint.

In summary, PTQ is the most reliable quantization technique for CPU inference: it accelerates all architectures except ShuffleNet-V2 and minimizes memory footprint in compact models. DQ is only competitive for large models like ResNet-152 and EfficientNet-B2, where its weight savings outweigh runtime overhead. 

\textbf{Early-Exit (EE)} behavior is governed by two key factors: the confidence threshold (acc-opt vs. inf-opt) and the exit depth, with architecture-dependent trade-offs.

For \textit{acc-opt}, prioritize fidelity over efficiency. For ResNet-152, speed-ups are marginal (0.98x–1.06x), as only 5–26\% of samples exit early—preserving near-baseline accuracy but offering negligible latency reduction. EfficientNet-B2 shows slight acceleration only in deeper exits (EE15: 1.09x), where features are sufficiently mature to enable confident early decisions without accuracy loss. For MobileNet-V2, \textit{acc-opt} configurations degrade performance (0.85x–0.95x) because early exits rarely trigger (<10\% exit rate), yet every sample incurs the overhead of passing through the additional classifier branches. ShuffleNet-V2 presents a more revealing case: despite achieving non-negligible exit rates (20–37\% in EE5/EE8), it still fails to accelerate (0.92x–0.97x). This confirms that in highly compact architectures, the computational cost of the early-exit classifier itself can outweigh the savings from skipping backbone layers, even when a substantial fraction of inputs exit early.

In the case of \textit{inf-opt}, reveal the true efficiency potential of early exits, with gains proportional to model size. ResNet-152 achieves the most substantial speed-ups (1.37x–2.24x), driven by exit rates of 79–99\%. Shallower exits (EE3) maximize savings, while deeper exits (EE14) trade lower acceleration (1.37x–1.40x) for better accuracy retention. EfficientNet-B2 follows a similar pattern with peak speed-ups of 1.52x–1.63x at intermediate exits (EE3/EE8), demonstrating that even moderately deep architectures benefit significantly from early branching. MobileNet-V2 shows diminishing returns: optimal speed-ups reach 1.24x–1.26x at shallow-to-intermediate exits (EE3/EE8), but gains shrink to 1.13x–1.17x when exits are placed too close to the final layer (EE13)—barely above pruning. ShuffleNet-V2 shows the most minor improvements (1.08x–1.15x), confirming that in highly compact architectures the computational cost of early-exit classifiers offsets the limited backbone savings, regardless of exit placement.

Examining Early-Exit / Final-Exit average inference times, a consistent pattern emerges: early exits are substantially faster than final exits (e.g., ResNet-152 EE3: 73.4ms vs. 269.3ms). However, samples that reach the final exit must traverse both the full backbone and the added classifier branches, often incurring higher latency than the baseline model. This penalty is most pronounced in compact architectures like ShuffleNet-V2, where final-exit inference time reaches 15.5ms on RPi5—52\% higher than the baseline (10.2ms). Deeper early-exit configurations (e.g., EE14) traverse more layers, limiting speed-ups compared to shallower exits, yet still outperform the baseline in total inference time.

CPU utilization reflects the efficiency patterns of each configuration. For \textit{acc-opt} thresholds, CPU usage remains near baseline across all devices for ResNet-152 (±0.1–0.3\%) and EfficientNet-B2 (±0.2–0.7\%). MobileNet-V2 shows slight increases (up to +2.8\%), while ShuffleNet-V2 exhibits the largest rises (up to +4.4\%), confirming that non-exiting samples still pay the overhead of traversing additional classifier branches. For \textit{inf-opt} thresholds, ResNet-152 achieves the clearest CPU reduction (e.g., Orin Nano: 98.3\% to 97.1\%), directly correlating with its high exit rates and substantial speed-ups. EfficientNet-B2 and MobileNet-V2 show smaller but consistent decreases (1.0–2.5\%). ShuffleNet-V2 again breaks the pattern: CPU usage either remains flat or increases (e.g., AGX Orin: 88.3\% to 89.9\%), confirming that in highly compact architectures, the computational cost of early-exit classifiers offsets the savings from skipped layers. GPU utilization under CUDA execution shows only marginal, device-dependent variations (±1–3\%) across both thresholds and all architectures, as early-exit branches execute primarily on CPU and the GPU workload remains dominated by the final backbone passes.

GPU utilization under CUDA execution shows distinct behavior between both configurations. For \textit{acc-opt}, GPU usage remains near baseline across all architectures, with minor fluctuations of ±1–3\% and no consistent trend—reflecting that accuracy-preserving thresholds trigger too few early exits to meaningfully alter the GPU workload. For \textit{inf-opt}, a clearer pattern emerges: GPU utilization consistently decreases in most configurations where speed-ups are significant, and the reduction is more pronounced with shallower early exits. ResNet-152 shows the clearest trend, with EE3 reducing GPU usage by 4–12\% (e.g., Orin Nano: 74.1\% to 62.7\%), while deeper EE14 shows smaller drops (74.1\% to 70.1\%). EfficientNet-B2 and MobileNet-V2 exhibit similar but more modest patterns (2–8\% reductions), with shallower exits consistently outperforming deeper ones. ShuffleNet-V2 again proves exceptional, with GPU usage either flat or slightly increasing—further evidence that its extreme compactness leaves insufficient computational depth for early exits to offload meaningful work from the GPU.

RAM footprint increases significantly with the addition of early-exit branches. Across all architectures and devices, acc-opt and inf-opt configurations consistently consume more memory than the baseline, with the penalty growing proportionally to the number and depth of exits.

In summary, early exits accelerate inference only in large models (ResNet-152: 1.37–2.24x), with diminishing returns in compact architectures, where classifier overhead often outweighs savings in the main structure. CPU utilization decreases significantly only when speed increases are achieved; the GPU's impact is minimal. All configurations increase RAM consumption proportionally to the number of exits.

\textbf{Overall}, Overall, quantization delivers the best compression and RAM savings (3.5–4x, up to 30–40\% memory reduction), with pruning also providing consistent but more modest memory gains (2–30\%). Early exits are the only technique that increases RAM, with overhead proportional to exit count. In terms of inference acceleration, early exits—when configured adequately with \textit{inf-opt} thresholds—achieve the highest peak speed-ups across all architectures, surpassing even quantization on large models (ResNet-152: 2.24x vs. \gls{ptq} 2.12x) and delivering measurable, albeit small, gains even on highly compact models like ShuffleNet-V2 (1.08–1.15x). \gls{ptq} offers the most consistent and hardware-friendly acceleration across the architectural spectrum (1.17–2.12x), while pruning provides reliable but more modest speed-ups (1.1–2.53x), competitive only on CPUs and large models. CPU and GPU utilization follow these trends: early exits and quantization reduce CPU load where they accelerate; pruning's impact is marginal. In short: quantization improves memory efficiency, early exits reduce latency for large models, and pruning remains the safest fallback.

\subsection{Early-Exit and Quantization Combination Analysis}
In this section, we extend the previous analysis by combining early exits and quantization under the same experimental setup and evaluation criteria used for the standalone optimization methods. The objective is to assess whether combining both techniques leads to additional benefits or introduces new trade-offs compared to applying early exits or quantization independently. The quantized models in this study (\gls{ptq} and \gls{dq}) were generated using an ONNX conversion pipeline that produces operators compatible with CPU execution engines. Consequently, all latency results reported in this section for quantized and quantized-EE models were measured on the CPU of the Jetson Orin Nano, focusing the analysis on CPU-bound edge inference scenarios.

This analysis uses the same early-exit configurations and confidence-threshold strategies (\textit{acc-opt} and \textit{inf-opt}) introduced in Section~\ref{sec:methodology}. For this comparison, we select the early-exit configuration that achieves the best accuracy-efficiency trade-off for each architecture: EE3 for ResNet-152, EE8 for EfficientNet-B2 and MobileNet-V2, and EE8 for ShuffleNet-V2. These configurations consistently provide the most favorable balance between accuracy retention and inference speed-up, as identified in Sections \ref{sec:results_A} and \ref{sec:results_B}. 

\begin{table*}[ht]
\caption{Early-Exit and Quantization Combination Across Architecture}
\label{tab:accuracy_shufflenet}
\begin{center}
\tiny
\begin{tabular}{|c|c|r|r|r|r|r|r|}
\hline
\textbf{Architecture} & \textbf{Model} &  \makecell{\textbf{Size} \\ \textbf{(MB)}} & \makecell{\textbf{Compression} \\ \textbf{Rate}} & \makecell{\textbf{Accuracy} \\ \textbf{(\%)}} & \makecell{\textbf{Total Inf.} \\ \textbf{Time (s)}} & \textbf{Speed-up} & \makecell{\textbf{Early-Exit} \\ \textbf{Rate (\%)}} \\

\hline
\multirow{9}{*}{\textbf{ResNet152}} & \textbf{BASE} & \textbf{233.20} & \textbf{1.00x} & \textbf{89.08} & \textbf{1327.4}  & \textbf{1.00x} &  -\\
\cline{2-8}
& PTQ & \textcolor{good}{59.10} & \textcolor{good}{3.95x} & 84.30 & 627.4 & 2.12x &  -\\
\cline{2-8}
& DQ & 59.20 & 3.94x & 88.96 & 736.2 & 1.80x &  -\\
\cline{2-8}
& EE3 (\textit{acc-opt}) & \multirow{2}{*}{\textcolor{bad}{233.60}} & \multirow{2}{*}{\textcolor{bad}{0.99x}} & \textcolor{good}{89.08} & \textcolor{bad}{1332.1} & \textcolor{bad}{1.08x} &  5.66\\
& EE3 (\textit{inf-opt}) &  &  & \textcolor{bad}{70.86} & 615.8 & 1.69x &  79.30\\
\cline{2-8}
& PTQ-EE3 (\textit{acc-opt}, T = 0.001) & \multirow{2}{*}{59.20} & \multirow{2}{*}{3.94x} & 88.68 & 513.0 & 2.59x &  2.12\\
& PTQ-EE3 (\textit{inf-opt}, T = 0.2 ) &  &  & 87.28 & \textcolor{good}{425.0} & \textcolor{good}{3.12x} & 29.64 \\
\cline{2-8}
& DQ-EE3 (\textit{acc-opt}, T = 0.011) & \multirow{2}{*}{\textcolor{good}{59.10}}& \multirow{2}{*}{\textcolor{good}{3.95x}} & 88.96 & 758.3 & 1.75x &  5.88\\
& DQ-EE3 (\textit{inf-opt}, T = 0.3) &  &  & 84.20 & 576.5 & 2.30x & 41.78 \\
\hline
\hline
\multirow{9}{*}{\textbf{EfficientNet-B2}} & \textbf{BASE} & \textbf{31.30} & \textbf{1.00x} & \textbf{91.18} & \textbf{402.4}  & \textbf{1.00x} &  -\\
\cline{2-8}
& PTQ & 9.00 & 3.47x & 64.86 & 273.3 & 1.47x &  -\\
\cline{2-8}
& DQ & \textcolor{good}{8.40} & \textcolor{good}{3.73x} & 11.08 & 522.9 & 0.77x &  -\\
\cline{2-8}
& EE8 (\textit{acc-opt}) & \multirow{2}{*}{\textcolor{bad}{31.53}} & \multirow{2}{*}{\textcolor{bad}{0.99x}} & \textcolor{good}{91.20} & 423.1 & 0.95x & 0.08 \\
& EE8 (\textit{inf-opt}) &  &  & 77.78 & 264.9 & 1.52x & 60.48 \\
\cline{2-8}
& PTQ-EE8 (\textit{acc-opt}, T = 0.001) & \multirow{2}{*}{8.64} & \multirow{2}{*}{3.62x} & 84.94 & 257.4 & 1.56x & 0.01 \\
& PTQ-EE8 (\textit{inf-opt}, T = 0.05) &  &  & 77.86 & \textcolor{good}{226.4} & \textcolor{good}{1.78x} & 0.10 \\
\cline{2-8}
& DQ-EE8 (\textit{acc-opt}, T = 0.001) & \multirow{2}{*}{8.43} & \multirow{2}{*}{3.71x} & 10.90 & \textcolor{bad}{542.7} & \textcolor{bad}{0.74x} & 0.05 \\
& DQ-EE8 (\textit{inf-opt}, T = 0.4) &  &  & \textcolor{bad}{4.70} & 297.7 & 1.35x & 0.84 \\
\hline
\hline
\multirow{9}{*}{\textbf{MobileNet-V2}} & \textbf{BASE} & \textbf{9.40} & \textbf{1.00x} & \textbf{87.80} & \textbf{179.1}  & \textbf{1.00x} &  -\\
\cline{2-8}
& PTQ & 2.60 & 3.62x & 81.22 & 97.0 & 1.85x &  -\\
\cline{2-8}
& DQ & \textcolor{good}{2.50} & \textcolor{good}{3.76x} & 86.82 & 234.7 & 0.76x &  -\\
\cline{2-8}
& EE8 (\textit{acc-opt}) & \multirow{2}{*}{\textcolor{bad}{9.58}} & \multirow{2}{*}{\textcolor{bad}{0.98x}} & \textcolor{good}{87.90} & 202.1 & 0.89x & 8.58\\
& EE8 (\textit{inf-opt}) &  &  & 70.10 & 147.1 & 1.22x & 73.64\\
\cline{2-8}
& PTQ-EE8 (\textit{acc-opt}, T = 0.05) & \multirow{2}{*}{2.59} & \multirow{2}{*}{3.63x} & 86.84 & 105.4 & 1.70x & 2.58 \\
& PTQ-EE8 (\textit{inf-opt}, T = 0.5) &  &  & 81.18 & \textcolor{good}{95.4} & \textcolor{good}{1.88x} & 43.08 \\
\cline{2-8}
& DQ-EE8 (\textit{acc-opt}, T = 0.04) & \multirow{2}{*}{2.62} & \multirow{2}{*}{3.59x} & 85.10 & \textcolor{bad}{254.9} & \textcolor{bad}{0.70x}& 1.68  \\
& DQ-EE8 (\textit{inf-opt}, T = 0.7) &  &  & \textcolor{bad}{49.78} & 172.9 & 1.04x & 82.60 \\
\hline
\hline
\multirow{9}{*}{\textbf{ShuffleNet-V2}} & \textbf{BASE} & \textbf{5.50} & \textbf{1.00x} & \textbf{77.86} & \textbf{99.6}  & \textbf{1.00x} &  -\\
\cline{2-8}
& PTQ & \textcolor{good}{1.50} & \textcolor{good}{3.67x} & 76.84 & \textcolor{bad}{152.0} & \textcolor{bad}{0.66x} &  -\\
\cline{2-8}
& DQ & \textcolor{good}{1.50} & \textcolor{good}{3.67x} & 75.74 & 121.5 & 0.82x &  -\\
\cline{2-8}
& EE8 (\textit{acc-opt}) & \multirow{2}{*}{\textcolor{bad}{5.90}} & \multirow{2}{*}{\textcolor{bad}{0.93x}} & \textcolor{good}{78.18} & 102.5 & 0.97x &  0.10\\
& EE8 (\textit{inf-opt}) &  &  & 73.72 & 91.9 & 1.08x & 38.84 \\
\cline{2-8}
& PTQ-EE8 (\textit{acc-opt}, T = 0.1) & \multirow{2}{*}{1.85} & \multirow{2}{*}{2.98x}  & 77.54 & 97.2 & 1.02x &  35.62\\
& PTQ-EE8 (\textit{inf-opt}, T = 0.5) &  &  & 73.90 & \textcolor{good}{88.4} & \textcolor{good}{1.30x} & 68.98 \\
\cline{2-8}
& DQ-EE8 (\textit{acc-opt}, T = 0.005) & \multirow{2}{*}{1.88} & \multirow{2}{*}{2.92x} & 75.60 & 133.6 & 0.74x & 13.80 \\
& DQ-EE8 (\textit{inf-opt}, T = 0.6) &  &  & \textcolor{bad}{69.68} & 110.7 & 0.75x &  66.28\\
\hline
\end{tabular}
\end{center}
\end{table*}

The key difference lies in how quantization is applied: instead of quantizing the original backbone network, we apply quantization after early-exit insertion—i.e., to the partitioned models generated by the early-exit design, as described in Section~\ref{sec:methodology_exp}. Each inference path (one per early exit, plus the final exit) is thus quantized independently using either \gls{ptq} or \gls{dq}.

The results, reported in Table~\ref{tab:accuracy_shufflenet}, show that combining early exits with quantization introduces systematically different trade-offs compared to applying each optimization independently.

\textbf{\gls{ptq}-EE} consistently outperforms \gls{ptq}-alone across all architectures — and crucially, it inherits the full compression benefit of quantization, maintaining model size reductions of 3x-4x. For ResNet-152, \gls{ptq}-EE3 inf-opt achieves 3.12x speed-up — substantially higher than \gls{ptq}-alone (2.12x) and even EE-alone (1.69x) — while recovering 87.28\% accuracy, a +2.98\% gain over \gls{ptq}-alone (84.30\%). This improvement comes from two sources: early exits reduce the fraction of samples traversing the full quantized backbone, and independent per-path quantization limits error accumulation in deep layers. For EfficientNet-B2, \gls{ptq}-EE8 inf-opt delivers 1.78x speed-up (vs. \gls{ptq}: 1.47x) and 77.86\% accuracy — a +13.0\% gain over \gls{ptq}-alone (64.86\%), all while keeping the model 3.6x smaller than the baseline. MobileNet-V2 follows the same pattern and even ShuffleNet-V2, where \gls{ptq}-alone degrades latency (0.66x), benefits from \gls{ptq}-EE8 inf-opt: 1.30x speed-up and 73.90\% accuracy (vs. \gls{ptq}: 76.84\% but with severe slowdown). The accuracy gain is smaller here, but the technique turns a latency regression into meaningful acceleration — again, without sacrificing the compression. Crucially, these latency benefits are not limited to \textit{inf-opt} configurations.

The benefits are most pronounced in \textit{inf-opt} configurations, where early exits actively reduce the number of samples reaching deep quantized layers. In contrast, \textit{acc-opt} configurations show modest speed-ups but near-baseline accuracy — confirming that the combination is best leveraged for latency-sensitive deployment.

\textbf{\gls{dq}-EE} presents a fundamentally different picture. While \gls{dq}-alone already struggles in compact architectures, adding early exits does not reverse this trend — and compression gains are comparable to PTQ-EE, but at the cost of catastrophic accuracy loss. \gls{dq}-EE8 \textit{inf-opt} on EfficientNet-B2 collapses to 4.70\% accuracy (vs. \gls{dq}: 11.08\%), and on MobileNet-V2 drops to 49.78\% (vs. \gls{dq}: 86.82\%). Speed-ups are either modest (1.04x–1.35x) or negative. The only viable case is ResNet-152, where \gls{dq}-EE3 inf-opt achieves 2.30x speed-up and 84.20\% accuracy — a respectable trade-off, though still inferior to \gls{ptq}-EE (3.12x, 87.28\%). This confirms that \gls{dq}'s runtime dynamic scaling compounds poorly with early-exit branching, amplifying quantization noise and destabilizing confidence estimates across multiple exit paths.

\textbf{In summary}, \gls{ptq}-EE is a highly effective hybrid optimization: it inherits \gls{ptq}'s compression and hardware compatibility while leveraging early exits to mitigate accuracy loss and boost speed-ups beyond what either technique achieves alone. \gls{dq}-EE, by contrast, is viable only for large models and remains inferior to \gls{ptq}-EE in all scenarios.

\section{Conclusions and Future Work}\label{sec:conclusions}
This study systematically analyzes the impact of various model optimization techniques for edge deployment, emphasizing the trade-offs between accuracy and inference latency under resource constraints. 

Architecture-specific results reveal no single optimal technique, but rather distinct trade-offs that depend on both the model class and the target metric. Large models (e.g., ResNet-152) accommodate all methods effectively: early exits deliver the highest inference speed-ups, while quantization dominates static compression. All techniques preserve accuracy within ±2\% of the baseline — with the sole exception of \textit{inf-opt} early exits, which trade greater latency reduction for a more pronounced accuracy penalty.

For quantization-sensitive architectures — exemplified by EfficientNet-B2 — pruning remains the safest compression alternative. However, even in these models, early exits with \textit{inf-opt} thresholds achieve substantial latency reductions, making them attractive for throughput-oriented server deployment despite notable accuracy degradation.

Compact edge-oriented models (e.g., MobileNet-V2, ShuffleNet-V2) present a different landscape: quantization remains the best option for static compression and memory savings — critical for on-device storage. Yet early exits become increasingly competitive at the edge: when placed at intermediate layers to avoid sharp accuracy drops, they improve inference times across the board — occasionally outperforming quantization — while retaining accuracy close to other optimization methods. This makes early exits particularly compelling for battery-powered or real-time edge applications where latency matters as much as model footprint. Pruning, despite offering decent compression, delivers more modest inference gains in this segment due to conservative pruning ratios necessary to preserve accuracy. 

Notably, the combination of early exits with quantization is transformative. Even in quantization-sensitive architectures such as EfficientNet-B2, where standalone \gls{ptq} reduces to 64.86\% accuracy, equipping the quantized model with early-exit heads — using \textit{acc-opt} thresholds — restores accuracy to 84.94\% while simultaneously reducing inference time. This hybrid approach delivers the best of both worlds: the compact memory footprint of a quantized model and the latency benefits of early exits, with accuracy matching or exceeding the full-precision baseline. For edge deployment, where storage, battery, and real-time constraints collide, this synergy turns a traditional trade-off into a genuine win-win. It represents, without question, the most compelling path forward for efficient inference at the extreme edge.

Future work could explore combining early exits with structured pruning to jointly exploit dynamic layer skipping and static parameter removal, particularly in overparameterized architectures.

Another direction involves hardware-specific deployment across multiple backends (CPU with OpenVINO, GPU with TensorRT, NPU with TFLite), where quantization and pruning benefits require backend-aware assessment beyond framework-level metrics. Crucially, the impact of early-exit training strategies on deployment efficiency remains underexplored; the two-stage training approach adopted here \cite{Fernandez2025Cap, kubaty2025how, bakhtiarnia2021multi}, while effective, may not be optimal for resource-constrained edge scenarios. Further research is needed to investigate how different training schemes—such as exit-specific distillation, joint supervision, or curriculum learning—affect not only accuracy but also memory footprint and peak inference cost.

Finally, adaptive input-aware exit policies present a promising direction, enabling dynamic threshold adjustment based on latency budgets, energy constraints, or input complexity.

\bibliographystyle{IEEEtran}
\bibliography{main}

@ARTICLE{Zhou_Paving,
  author={Zhou, Zhi and Chen, Xu and Li, En and Zeng, Liekang and Luo, Ke and Zhang, Junshan},
  journal={Proceedings of the IEEE}, 
  title={Edge Intelligence: Paving the Last Mile of Artificial Intelligence With Edge Computing}, 
  year={2019},
  volume={107},
  number={8},
  pages={1738-1762},
  doi={10.1109/JPROC.2019.2918951}}

@article{Wang_Empowering,
author = {Wang, Xubin and Tang, Zhiqing and Guo, Jianxiong and Meng, Tianhui and Wang, Chenhao and Wang, Tian and Jia, Weijia},
title = {Empowering Edge Intelligence: A Comprehensive Survey on On-Device AI Models},
year = {2025},
issue_date = {September 2025},
publisher = {Association for Computing Machinery},
address = {New York, NY, USA},
volume = {57},
number = {9},
issn = {0360-0300},
url = {https://doi.org/10.1145/3724420},
doi = {10.1145/3724420},
journal = {ACM Comput. Surv.},
month = apr,
articleno = {228},
numpages = {39}
}

@article{Luiz_Integration,
title = {The Internet of Things, Fog and Cloud continuum: Integration and challenges},
journal = {Internet of Things},
volume = {3-4},
pages = {134-155},
year = {2018},
issn = {2542-6605},
doi = {https://doi.org/10.1016/j.iot.2018.09.005},
url = {https://www.sciencedirect.com/science/article/pii/S2542660518300635},
author = {Luiz Bittencourt and Roger Immich and Rizos Sakellariou and Nelson Fonseca and Edmundo Madeira and Marilia Curado and Leandro Villas and Luiz DaSilva and Craig Lee and Omer Rana}
}

@InCollection{Daghero_Energy-efficient,
  author           = {Francesco Daghero and Daniele Jahier Pagliari and Massimo Poncino},
  booktitle        = {Hardware Accelerator Systems for Artificial Intelligence and Machine Learning},
  publisher        = {Elsevier},
  title            = {Chapter Eight - Energy-efficient deep learning inference on edge devices},
  year             = {2021},
  editor           = {Shiho Kim and Ganesh Chandra Deka},
  pages            = {247-301},
  series           = {Advances in Computers},
  volume           = {122},
  creationdate     = {2025-06-22T17:20:43},
  doi              = {https://doi.org/10.1016/bs.adcom.2020.07.002},
  issn             = {0065-2458},
  modificationdate = {2025-06-22T17:20:46},
  url              = {https://www.sciencedirect.com/science/article/pii/S0065245820300553},
}

@article{Kartikeya_Vision,
  author       = {Kartikeya Bhardwaj and
                  Naveen Suda and
                  Radu Marculescu},
  title        = {{EdgeAI}: {A} Vision for Deep Learning in {IoT} Era},
  journal      = {CoRR},
  volume       = {abs/1910.10356},
  year         = {2019},
  url          = {http://arxiv.org/abs/1910.10356},
  eprinttype    = {arXiv},
  eprint       = {1910.10356},
  bibsource    = {dblp computer science bibliography, https://dblp.org}
}

@Article{Satyanarayanan_Emergence,
  author           = {Satyanarayanan, Mahadev},
  journal          = {Computer},
  title            = {The Emergence of Edge Computing},
  year             = {2017},
  number           = {1},
  pages            = {30-39},
  volume           = {50},
  creationdate     = {2025-06-22T17:28:56},
  doi              = {10.1109/MC.2017.9},
  modificationdate = {2025-06-22T17:28:58},
}

@Article{Dantas_Comprehensive,
  author           = {Dantas, Pierre Vilar and Sabino da Silva, Waldir and Cordeiro, Lucas Carvalho and Carvalho, Celso Barbosa},
  journal          = {Applied Intelligence},
  title            = {A comprehensive review of model compression techniques in machine learning},
  year             = {2024},
  issn             = {1573-7497},
  month            = nov,
  number           = {22},
  pages            = {11804--11844},
  volume           = {54}
}

@InProceedings{Kuzmin_prunin,
  author           = {Kuzmin, Andrey and Nagel, Markus and Van Baalen, Mart and Behboodi, Arash and Blankevoort, Tijmen},
  title            = {Pruning vs Quantization: Which is Better?},
  booktitle = {Advances in Neural Information Processing Systems},
   year             = {2024},
  month            = {2}}

@InProceedings{Agrawal_Efficient,
  author           = {Agrawal, Rishabh and Kumar, Himanshu and Lnu, Shashikant Reddy},
  title            = {Efficient {LLMs} for {Edge} {Devices}: {Pruning}, {Quantization}, and {Distillation} {Techniques}},
  year             = {2025},
  booktitle={2025 International Conference on Machine Learning and Autonomous Systems (ICMLAS)},
  month            = mar,
  pages            = {1413--1418}}

@incollection{Fernandez2025Cap,
  author    = {Fernandez, Nekane and Amurrio, Andoni and Van Vaerenbergh, Steven},
  title     = {{Optimization Approaches for Distributed AI Models on Edge Devices}},
  booktitle = {Novel Deep Learning Methodologies in Industrial and Applied Mathematics},
  editor    = {S. Xambó-Descamps},
  year      = {2025},
  publisher = {Springer Nature},
  note      = {Proceedings of the ICIAM MS 02515 organized within the ICIAM-2023 Conference, in press}
}

@inproceedings{teerapittayanon2016branchynet,
  title={Branchynet: Fast inference via early exiting from deep neural networks},
  author={Teerapittayanon, Surat and McDanel, Bradley and Kung, Hsiang-Tsung},
  booktitle={2016 23rd international conference on pattern recognition (ICPR)},
  pages={2464--2469},
  year={2016},
  organization={IEEE}
}

@Article{Li_model_compression,
  author           = {Li, Zhuo and Li, Hengyi and Meng, Lin},
  journal          = {Computers},
  title            = {Model Compression for Deep Neural Networks: A Survey},
  year             = {2023},
  issn             = {2073-431X},
  number           = {3},
  volume           = {12},
  article-number   = {60},
  creationdate     = {2025-06-25T08:02:14},
  doi              = {10.3390/computers12030060},
  modificationdate = {2025-06-25T08:02:14},
  url              = {https://www.mdpi.com/2073-431X/12/3/60},
}

@InProceedings{Yan_Polythrottle,
  author           = {Yan, Minghao and Wang, Hongyi and Venkataraman, Shivaram},
  title            = {PolyThrottle: Energy-efficient Neural Network Inference on Edge Devices},
  year             = {2023},
  creationdate     = {2025-06-25T12:13:52},
  modificationdate = {2025-06-25T12:13:52},
}

@Article{Xu_EfficientHardware,
  author           = {Xu, Xiaowei and Lu, Qing and Wang, Tianchen and Hu, Yu and Zhuo, Chen and Liu, Jinglan and Shi, Yiyu},
  journal          = {J. Emerg. Technol. Comput. Syst.},
  title            = {Efficient Hardware Implementation of Cellular Neural Networks with Incremental Quantization and Early Exit},
  year             = {2018},
  issn             = {1550-4832},
  month            = dec,
  number           = {4},
  volume           = {14},
  address          = {New York, NY, USA},
  articleno        = {48},
  creationdate     = {2025-06-25T12:17:05},
  doi              = {10.1145/3264817},
  issue_date       = {October 2018},
  modificationdate = {2025-06-25T12:17:08},
  numpages         = {20}}

@InProceedings{Korol_Pruning,
  author           = {Korol, Guilherme and Jordan, Michael Guilherme and Rutzig, Mateus Beck and Castrillon, Jeronimo and Beck, Antonio Carlos Schneider},
  booktitle        = {2023 Design, Automation \& Test in Europe Conference \& Exhibition (DATE)},
  title            = {Pruning and Early-Exit Co-Optimization for CNN Acceleration on FPGAs},
  year             = {2023},
  pages            = {1-6},
  creationdate     = {2025-06-25T12:20:37},
  doi              = {10.23919/DATE56975.2023.10137244},
  modificationdate = {2025-06-25T12:20:39},
}

@misc{Piveral_Memory,
  author = {Piveral},
  title = {Memory Architecture and CUDA Programming on Jetson Orin: Differences from x86 {GPUs} - Help Docs for Errors/Issues on Nvidia Jetson Dev Boards},
  year = {2024},
  howpublished = {\url{https://nvidia-jetson.piveral.com/jetson-orin-nano/memory-architecture-and-cuda-programming-on-jetson-orin-differences-from-x86-gpus}},
  note = {Accedido el \today}
}

@article{Liu2017LearningEC,
  title={Learning Efficient Convolutional Networks through Network Slimming},
  author={Zhuang Liu and Jianguo Li and Zhiqiang Shen and Gao Huang and Shoumeng Yan and Changshui Zhang},
  journal={2017 IEEE International Conference on Computer Vision (ICCV)},
  year={2017},
  pages={2755-2763},
  url={https://api.semanticscholar.org/CorpusID:5993328}
}

@article{DBLP:journals/corr/abs-1810-07052,
  author       = {Yigitcan Kaya and
                  Tudor Dumitras},
  title        = {How to Stop Off-the-Shelf Deep Neural Networks from Overthinking},
  journal      = {CoRR},
  volume       = {abs/1810.07052},
  year         = {2018},
  url          = {http://arxiv.org/abs/1810.07052},
  eprinttype    = {arXiv},
  eprint       = {1810.07052},
  bibsource    = {dblp computer science bibliography, https://dblp.org}
}

@Article{Vaiyapuri2025,
  author           = {Vaiyapuri, Thavavel and Aldosari, Huda},
  journal          = {Sustainability},
  title            = {{SUQ-3}: A Three Stage Coarse-to-Fine Compression Framework for Sustainable Edge {AI} in Smart Farming},
  year             = {2025},
  issn             = {2071-1050},
  month            = jun,
  number           = {12},
  pages            = {5230},
  volume           = {17},
  creationdate     = {2025-06-19T16:29:38},
  date             = {2025-06-06},
  day              = {6},
  doi              = {10.3390/su17125230},
  editor           = {Lucia Rocchi},
  modificationdate = {2025-06-25T08:29:38},
  publisher        = {MDPI AG},
}

@misc{francy2024edgeaievaluationmodel,
      title={Edge AI: Evaluation of Model Compression Techniques for Convolutional Neural Networks}, 
      author={Samer Francy and Raghubir Singh},
      year={2024},
      eprint={2409.02134},
      archivePrefix={arXiv},
      primaryClass={cs.LG},
      url={https://arxiv.org/abs/2409.02134}, 
}

@Article{Paranayapa2024,
  author           = {Paranayapa, Thivindu and Ranasinghe, Piumini and Ranmal, Dakshina and Meedeniya, Dulani and Perera, Charith},
  journal          = {Sensors},
  title            = {A Comparative Study of Preprocessing and Model Compression Techniques in Deep Learning for Forest Sound Classification},
  year             = {2024},
  issn             = {1424-8220},
  month            = feb,
  number           = {4},
  pages            = {1149},
  volume           = {24}}

@Article{Karathanasis2025,
  author           = {Karathanasis, Andreas and Violos, John and Kompatsiaris, Ioannis},
  journal          = {Mathematics},
  title            = {A Comparative Analysis of Compression and Transfer Learning Techniques in DeepFake Detection Models},
  year             = {2025},
  issn             = {2227-7390},
  number           = {5},
  volume           = {13},
  article-number   = {887},
  creationdate     = {2025-07-24T12:54:37},
  doi              = {10.3390/math13050887},
  modificationdate = {2025-07-24T12:54:43},
  url              = {https://www.mdpi.com/2227-7390/13/5/887},
}

@article{Naveen2024OptimizedCN,
  title={Optimized Convolutional Neural Network at the {IoT} edge for image detection using pruning and quantization},
  author={Soumyalatha Naveen and Manjunath R. Kounte},
  journal={Multim. Tools Appl.},
  year={2024},
  volume={84},
  pages={5435-5455},
  url={https://api.semanticscholar.org/CorpusID:275070866}
}

@InProceedings{Wang_2018_ECCV,
author = {Wang, Xin and Yu, Fisher and Dou, Zi-Yi and Darrell, Trevor and Gonzalez, Joseph E.},
title = {SkipNet: Learning Dynamic Routing in Convolutional Networks},
booktitle = {Proceedings of the European Conference on Computer Vision (ECCV)},
month = {September},
year = {2018}
}

@inproceedings{10.5555/3540261.3541329,
author = {Rao, Yongming and Zhao, Wenliang and Liu, Benlin and Lu, Jiwen and Zhou, Jie and Hsieh, Cho-Jui},
title = {DynamicViT: efficient vision transformers with dynamic token sparsification},
year = {2021},
isbn = {9781713845393},
publisher = {Curran Associates Inc.},
address = {Red Hook, NY, USA},
booktitle = {Proceedings of the 35th International Conference on Neural Information Processing Systems},
articleno = {1068},
numpages = {13},
series = {NIPS '21}
}

@article{mohanty2025pruning,
  title={Pruning techniques for artificial intelligence networks: a deeper look at their engineering design and bias: the first review of its kind},
  author={Mohanty, Lopamudra and Kumar, Ashish and Mehta, Vivek and Agarwal, Mohit and Suri, Jasjit S},
  journal={Multimedia Tools and Applications},
  volume={84},
  number={11},
  pages={9591--9665},
  year={2025},
  publisher={Springer}
}

@inproceedings{wang2020sparsert,
  title={Sparsert: Accelerating unstructured sparsity on gpus for deep learning inference},
  author={Wang, Ziheng},
  booktitle={Proceedings of the ACM international conference on parallel architectures and compilation techniques},
  pages={31--42},
  year={2020}
}

@inproceedings{yao2019balanced,
  title={Balanced sparsity for efficient dnn inference on gpu},
  author={Yao, Zhuliang and Cao, Shijie and Xiao, Wencong and Zhang, Chen and Nie, Lanshun},
  booktitle={Proceedings of the AAAI conference on artificial intelligence},
  volume={33},
  number={01},
  pages={5676--5683},
  year={2019}
}

@inproceedings{belhadi2025lightprune,
  title={LightPrune: Latency-Aware Structured Pruning for Efficient Deep Inference on Embedded Devices},
  author={Belhadi, Asma and Djenouri, Youcef and Belbachir, Ahmed Nabil},
  booktitle={Proceedings of the IEEE/CVF International Conference on Computer Vision},
  pages={1688--1697},
  year={2025}
}

@article{XU2025122470,
title = {{DEEP-CWS}: Distilling Efficient pre-trained models with Early exit and Pruning for scalable Chinese Word Segmentation},
journal = {Information Sciences},
volume = {719},
pages = {122470},
year = {2025},
issn = {0020-0255},
doi = {https://doi.org/10.1016/j.ins.2025.122470},
url = {https://www.sciencedirect.com/science/article/pii/S0020025525006024},
author = {Shiting, Xu}
}

@inproceedings{10.1145/3587135.3592204,
author = {Ghanathe, Nikhil P. and Wilton, Steve},
title = {T-RecX: Tiny-Resource Efficient Convolutional neural networks with early-eXit},
year = {2023},
isbn = {9798400701405},
publisher = {Association for Computing Machinery},
address = {New York, NY, USA},
url = {https://doi.org/10.1145/3587135.3592204},
doi = {10.1145/3587135.3592204},
booktitle = {Proceedings of the 20th ACM International Conference on Computing Frontiers},
pages = {123–133},
numpages = {11},
location = {Bologna, Italy},
series = {CF '23}
}

@ARTICLE{10929047,
  author={U, Vivek Menon and Babu Kumaravelu, Vinoth and C, Vinoth Kumar and A, Rammohan and Chinnadurai, Sunil and Venkatesan, Rajeshkumar and Hai, Han and Selvaprabhu, Poongundran},
  journal={IEEE Access}, 
  title={{AI}-Powered {IoT}: A Survey on Integrating Artificial Intelligence With {IoT} for Enhanced Security, Efficiency, and Smart Applications}, 
  year={2025},
  volume={13},
  number={},
  pages={50296-50339},
  doi={10.1109/ACCESS.2025.3551750}}

@article{yuan2023usdc,
  title={USDC: Unified Static and Dynamic Compression for Visual Transformer},
  author={Yuan, Huan and Liao, Chao and Tan, Jianchao and Yao, Peng and Jia, Jiyuan and Chen, Bin and Song, Chengru and Zhang, Di},
  journal={arXiv preprint arXiv:2310.11117},
  year={2023}
}

@ARTICLE{9560049,
  author={Han, Yizeng and Huang, Gao and Song, Shiji and Yang, Le and Wang, Honghui and Wang, Yulin},
  journal={IEEE Transactions on Pattern Analysis and Machine Intelligence}, 
  title={Dynamic Neural Networks: A Survey}, 
  year={2022},
  volume={44},
  number={11},
  pages={7436-7456},
  doi={10.1109/TPAMI.2021.3117837}}

@article{krishnamoorthi2018quantizing,
  title={Quantizing deep convolutional networks for efficient inference: A whitepaper},
  author={Krishnamoorthi, Raghuraman},
  journal={arXiv preprint arXiv:1806.08342},
  year={2018}
}

@inproceedings{zhao2024streamlining,
  title={Streamlining Speech Enhancement DNNs: an Automated Pruning Method Based on Dependency Graph with Advanced Regularized Loss Strategies},
  author={Zhao, Zugang and Zhang, Jinghong and Liu, Yonghui and Liu, Jianbing and Niu, Kai and He, Zhiqiang},
  booktitle={Proc. Interspeech 2024},
  pages={662--666},
  year={2024}
}

@article{DBLP:journals/corr/abs-2004-09576,
  author       = {Yash Bhalgat and
                  Jinwon Lee and
                  Markus Nagel and
                  Tijmen Blankevoort and
                  Nojun Kwak},
  title        = {{LSQ+:} Improving low-bit quantization through learnable offsets and
                  better initialization},
  journal      = {CoRR},
  volume       = {abs/2004.09576},
  year         = {2020},
  url          = {https://arxiv.org/abs/2004.09576},
  eprinttype    = {arXiv},
  eprint       = {2004.09576},
  bibsource    = {dblp computer science bibliography, https://dblp.org}
}

@article{DBLP:journals/corr/abs-1802-10280,
  author       = {Xuhao Chen},
  title        = {Escort: Efficient Sparse Convolutional Neural Networks on GPUs},
  journal      = {CoRR},
  volume       = {abs/1802.10280},
  year         = {2018},
  url          = {http://arxiv.org/abs/1802.10280},
  eprinttype    = {arXiv},
  eprint       = {1802.10280},
  bibsource    = {dblp computer science bibliography, https://dblp.org}
}

@Article{electronics14173468,
    AUTHOR = {Huang, Xiaoyuan and Wang, Hongcheng and Qin, Shiyin and Tang, Su-Kit},
    TITLE = {Embedded Artificial Intelligence: A Comprehensive Literature Review},
    JOURNAL = {Electronics},
    VOLUME = {14},
    YEAR = {2025},
    NUMBER = {17},
    ARTICLE-NUMBER = {3468},
    URL = {https://www.mdpi.com/2079-9292/14/17/3468},
    ISSN = {2079-9292},
    DOI = {10.3390/electronics14173468}
}

@article{kumar2021pruning,
  title={Pruning filters with L1-norm and capped L1-norm for CNN compression},
  author={Kumar, Aakash and Shaikh, Ali Muhammad and Li, Yun and Bilal, Hazrat and Yin, Baoqun},
  journal={Applied Intelligence},
  volume={51},
  number={2},
  pages={1152--1160},
  year={2021},
  publisher={Springer}
}

@inproceedings{10.1145/3650212.3680374,
author = {Jajal, Purvish and Jiang, Wenxin and Tewari, Arav and Kocinare, Erik and Woo, Joseph and Sarraf, Anusha and Lu, Yung-Hsiang and Thiruvathukal, George K. and Davis, James C.},
title = {Interoperability in Deep Learning: A User Survey and Failure Analysis of ONNX Model Converters},
year = {2024},
isbn = {9798400706127},
publisher = {Association for Computing Machinery},
address = {New York, NY, USA},
url = {https://doi.org/10.1145/3650212.3680374},
doi = {10.1145/3650212.3680374},
booktitle = {Proceedings of the 33rd ACM SIGSOFT International Symposium on Software Testing and Analysis},
pages = {1466–1478},
numpages = {13},
location = {Vienna, Austria},
series = {ISSTA 2024}
}

@article{nagel2021white,
  title={A white paper on neural network quantization},
  author={Nagel, Markus and Fournarakis, Marios and Amjad, Rana Ali and Bondarenko, Yelysei and Van Baalen, Mart and Blankevoort, Tijmen},
  journal={arXiv preprint arXiv:2106.08295},
  year={2021}
}

@misc{brevitas2021,
  author = {Pappalardo, A. and others},
  title  = {{Xilinx/Brevitas: v0.7.1}},
  year   = {2021},
  note   = {[Online]. Available: \url{https://github.com/Xilinx/brevitas}}
}

@inproceedings{kubaty2025how,
title={How to Train Your Multi-Exit Model? Analyzing the Impact of Training Strategies},
author={Piotr Kubaty and Bartosz W{\'o}jcik and Bart{\l}omiej Tomasz Krzepkowski and Monika Michaluk and Tomasz Trzcinski and Jary Pomponi and Kamil Adamczewski},
booktitle={Forty-second International Conference on Machine Learning},
year={2025},
url={https://openreview.net/forum?id=vhTPfOdwyQ}
}

@article{bakhtiarnia2021multi,
  title={Multi-exit vision transformer for dynamic inference},
  author={Bakhtiarnia, Arian and Zhang, Qi and Iosifidis, Alexandros},
  journal={arXiv preprint arXiv:2106.15183},
  year={2021}
}

@misc{pytorch_torch_cond,
  author       = {{PyTorch Team}},
  title        = {torch.cond},
  howpublished = {\emph{PyTorch Documentation}},
  year         = {n.d.},
  note         = {[Online]. Available: https://docs.pytorch.org/docs/stable/generated/torch.cond.html}
}

@article{zhao2022probability,
  author    = {Zhao, H. L. and Shi, K. J. and Jin, X. G. and others},
  title     = {Probability-Based Channel Pruning for Depthwise Separable Convolutional Networks},
  journal   = {Journal of Computer Science and Technology},
  volume    = {37},
  number    = {3},
  pages     = {584--600},
  year      = {2022},
  doi       = {10.1007/s11390-022-2131-8},
  url       = {https://doi.org/10.1007/s11390-022-2131-8}
}

@article{Gong2023Dynamic,
title={Dynamic Shuffle: An Efficient Channel Mixture Method},
author={Kaijun Gong and Zhuowen Yin and Yushu Li and K. Guo and Xiangmin Xu},
journal={ArXiv},
year={2023},
volume={abs/2310.02776},
doi={10.48550/arxiv.2310.02776}
}

@article{Lin2024QServe:,
    title={QServe: W4A8KV4 Quantization and System Co-design for Efficient LLM Serving},
    author={Yujun Lin and Haotian Tang and Shang Yang and Zhekai Zhang and Guangxuan Xiao and Chuang Gan and Song Han},
    journal={ArXiv},
    year={2024},
    volume={abs/2405.04532},
    doi={10.48550/arxiv.2405.04532}}

@article{Kim2024QUICK:,
    title={QUICK: Quantization-aware Interleaving and Conflict-free Kernel for efficient LLM inference},
    author={Taesu Kim and Jongho Lee and Daehyun Ahn and Sarang Kim and Jiwoong Choi and Minkyu Kim and Hyungjun Kim},
    journal={ArXiv},
    year={2024},
    volume={abs/2402.10076},
    doi={10.48550/arxiv.2402.10076}}

@INPROCEEDINGS{9042000,
  author={Radu, Valentin and Kaszyk, Kuba and Wen, Yuan and Turner, Jack and Cano, José and Crowley, Elliot J. and Franke, Björn and Storkey, Amos and O'Boyle, Michael},
  booktitle={2019 IEEE International Symposium on Workload Characterization (IISWC)}, 
  title={Performance Aware Convolutional Neural Network Channel Pruning for Embedded GPUs}, 
  year={2019},
  volume={},
  number={},
  pages={24-34},
  doi={10.1109/IISWC47752.2019.9042000}}

@article{LIU2021100009,
    title = {Latency-aware automatic CNN channel pruning with GPU runtime analysis},
    journal = {BenchCouncil Transactions on Benchmarks, Standards and Evaluations},
    volume = {1},
    number = {1},
    pages = {100009},
    year = {2021},
    issn = {2772-4859},
    doi = {https://doi.org/10.1016/j.tbench.2021.100009},
    url = {https://www.sciencedirect.com/science/article/pii/S2772485921000090},
    author = {Jiaqiang Liu and Jingwei Sun and Zhongtian Xu and Guangzhong Sun}
}

@article{KIM2025107718,
    title = {QuantuneV2: Compiler-based local metric-driven mixed precision quantization for practical embedded AI applications},
    journal = {Future Generation Computer Systems},
    volume = {166},
    pages = {107718},
    year = {2025},
    issn = {0167-739X},
    doi = {https://doi.org/10.1016/j.future.2025.107718},
    url = {https://www.sciencedirect.com/science/article/pii/S0167739X25000135},
    author = {Jeongseok Kim and Jemin Lee and Yongin Kwon and Daeyoung Kim}
}

@article{Semerikov2025Edge,
title={Edge intelligence unleashed: a survey on deploying large language models in resource-constrained environments},
author={Serhii Semerikov and T. Vakaliuk and O. Kanevska and O.A. Ostroushko and A. O. Kolhatin},
journal={J. Edge Comput.},
year={2025},
volume={4},
pages={179-233},
doi={10.55056/jec.1000}
}

@ARTICLE{10909993,
  author={Korol, Guilherme and Beck, Antonio Carlos Schneider},
  journal={IEEE Transactions on Very Large Scale Integration (VLSI) Systems}, 
  title={IoT-Edge Splitting With Pruned Early-Exit CNNs for Adaptive Inference}, 
  year={2025},
  volume={33},
  number={9},
  pages={2382-2394},
  doi={10.1109/TVLSI.2025.3543445}}

@inproceedings{saxena2023mcqueen,
  title={McQueen: Mixed Precision Quantization of Early Exit Networks.},
  author={Saxena, Utkarsh and Roy, Kaushik},
  booktitle={BMVC},
  pages={511--513},
  year={2023}
}

@incollection{deutel2026recent,
  title={Recent Trends in Edge {AI}: Efficient Design, Training and Deployment of Machine Learning Models},
  author={Deutel, Mark and Mallah, Maen and Wissing, Julio and Scheele, Stephan},
  booktitle={Charting the Intelligence Frontiers--Edge AI Systems Nexus},
  pages={181--220},
  year={2026},
  publisher={River Publishers}
}

@misc{psutil_cpu_percent,
  title        = {psutil.cpu\_percent — psutil 7.2.2 documentation},
  howpublished = {Online},
  note         = {[Online]. Available: \url{https://psutil.readthedocs.io/en/latest/\#psutil.cpu_percent}},
  year         = {2026},
  organization = {psutil development team},
  url          = {https://psutil.readthedocs.io/en/latest/#psutil.cpu_percent}
}

@misc{jetsonstats_gpu_class,
  title        = {jtop.core.gpu.GPU — jetson-stats 4.5.4 API Reference},
  howpublished = {Online},
  note         = {[Online]. Available: \url{https://rnext.it/jetson_stats/reference/gpu.html#jtop.core.gpu.GPU}},
  year         = {2025},
  organization = {jetson-stats development team},
  url          = {https://rnext.it/jetson_stats/reference/gpu.html#jtop.core.gpu.GPU}
}

@misc{jetsonstats_jtop_memory,
  title        = {jtop.jtop.memory — jetson-stats 4.5.4 API Reference},
  howpublished = {Online},
  note         = {[Online]. Available: \url{https://rnext.it/jetson_stats/reference/jtop.html#jtop.jtop.memory}},
  year         = {2025},
  organization = {jetson-stats development team},
  url          = {https://rnext.it/jetson_stats/reference/jtop.html#jtop.jtop.memory}
}

@misc{jain2020efficientexecutionquantizeddeep,
      title={Efficient Execution of Quantized Deep Learning Models: A Compiler Approach}, 
      author={Animesh Jain and Shoubhik Bhattacharya and Masahiro Masuda and Vin Sharma and Yida Wang},
      year={2020},
      eprint={2006.10226},
      archivePrefix={arXiv},
      primaryClass={cs.DC},
      url={https://arxiv.org/abs/2006.10226}, 
}

@article{Elhanashi2023Integration,
    title={Integration of Deep Learning into the IoT: A Survey of Techniques and Challenges for Real-World Applications},
    author={Abdussalam Elhanashi and Pierpaolo Dini and Sergio Saponara and Qinghe Zheng},
    journal={Electronics},
    year={2023},
    doi={10.3390/electronics12244925}
}

@article{Chen2020Deep,
    title={Deep Learning on Computational-Resource-Limited Platforms: A Survey},
    author={Chunlei Chen and Peng Zhang and Huixiang Zhang and Jiangyan Dai and Yugen Yi and Huihui Zhang and Yonghui Zhang},
    journal={Mob. Inf. Syst.},
    year={2020},
    volume={2020},
    pages={8454327:1-8454327:19},
    doi={10.1155/2020/8454327}
}

@ARTICLE{10398463,
  author={Cheng, Long and Gu, Yan and Liu, Qingzhi and Yang, Lei and Liu, Cheng and Wang, Ying},
  journal={IEEE Transactions on Sustainable Computing}, 
  title={Advancements in Accelerating Deep Neural Network Inference on AIoT Devices: A Survey}, 
  year={2024},
  volume={9},
  number={6},
  pages={830-847},
  doi={10.1109/TSUSC.2024.3353176}}

\appendix

\section{Appendix: Detailed Pruning Results}
\label{app:pruning}

Table~\ref{tab:pruning_configs} summarizes the results of multiple structured pruning configurations applied to different CNN architectures, including ResNet-152, EfficientNet-B2, and MobileNet-V2. For each model, we evaluated different combinations of channel granularity (\gls{cg}) and pruning ratio (\gls{pr})to analyze their impact on model size, compression ratio, and predictive performance.

The \gls{cg} parameter plays a critical role in determining pruning effectiveness. A \gls{cg} value of 16 operates at a finer granularity, allowing more precise removal of channels, which can preserve accuracy at low pruning ratios but often yields lower compression rates. Conversely, \gls{cg}=32 removes channels in coarser groups, achieving higher compression for the same pruning ratio—as seen consistently across architectures—but may lead to more abrupt capacity loss. The interaction between \gls{cg} and \gls{pr} is significant in compact models, where coarse-grained pruning at high ratios can rapidly degrade performance.

\begin{table*}[ht]
\caption{Results of different pruning configurations}
\label{tab:pruning_configs}
\centering
\tiny
\begin{tabular}{|c|c|r|r|r|r|r|}
\hline
\textbf{Architecture} & \textbf{Model} &  \makecell{\textbf{Size} \\ \textbf{(MB)}} & \makecell{\textbf{Compression} \\ \textbf{Rate}} & \makecell{\textbf{Accuracy} \\ \textbf{(\%)}} & \makecell{\textbf{Label Loyalty} \\ \textbf{(\%)}} & \makecell{\textbf{Prob. Loyalty} \\ \textbf{(\%)}} \\ 
\hline
\hline 
\multirow{9}{*}{\textbf{ResNet152}} & \textbf{BASE} & \textbf{233.2} & \textbf{1.00x} &  \textbf{89.08} & - & - \\
& CG=16, PR=0.05 & \textcolor{bad}{204.2} & \textcolor{bad}{1.14x} & 88.66 & 94.68 & 97.37 \\
& CG=32, PR=0.05 & 188.3 & 1.24x & 88.32 & 93.70 & 96.53 \\
& CG=16, PR=0.10 & 180.3 & 1.29x & \textcolor{good}{88.82} & \textcolor{good}{94.28} & \textcolor{good}{96.87} \\
& CG=32, PR=0.10 & 177.3 & 1.32x & 88.46 & 93.22 & 96.08 \\
& CG=16, PR=0.20 & 137.7 & 1.69x & 88.38 & 92.88 & 95.78 \\
& CG=32, PR=0.20 & 134.3 & 1.74x & 88.38 & 92.70 & 95.48 \\
& CG=16, PR=0.30 & 109.9 & 2.12x & 87.42 & 91.40 & 94.69 \\
& \textbf{\textit{CG=32, PR=0.30}} & \textbf{\textit{\textcolor{good}{98.5}}} & \textbf{\textit{\textcolor{good}{2.37x}}} & \textbf{\textit{\textcolor{bad}{87.22}}} & \textbf{\textit{\textcolor{bad}{90.88}}} & \textbf{\textit{\textcolor{bad}{94.11}}} \\
\hline
\hline
\multirow{9}{*}{\textbf{EfficientNet-B2}} & \textbf{BASE} & \textbf{31.3} & \textbf{1.00x} &  \textbf{91.18} & - & - \\
& CG=16, PR=0.05 & \textcolor{bad}{26.9} & \textcolor{bad}{1.16x} & 85.88 & 88.08 & 91.55 \\
& \textbf{\textit{CG=32, PR=0.05}} & \textbf{\textit{25.6}} & \textbf{\textit{1.20x}} & \textbf{\textit{\textcolor{good}{87.12}}} & \textbf{\textit{\textcolor{good}{89.48}}} & \textbf{\textit{\textcolor{good}{92.71}}}\\
& CG=16, PR=0.10 & 22.9 & 1.36x & 83.54 & 85.38 & 89.11\\
& CG=32, PR=0.10 & 22.2 & 1.41x & 86.54 & 88.48 & 91.80\\
& CG=16, PR=0.20 & 19.1 & 1.64x & 80.64 & 82.08 & 85.96\\
& CG=32, PR=0.20 & 19.0 & 1.65x & 83.58 & 85.58 & 89.19\\
& CG=16, PR=0.30 & 14.9 & 2.10x &  \textcolor{bad}{74.48} & \textcolor{bad}{75.94} & \textcolor{bad}{80.41} \\
& CG=32, PR=0.30 & \textcolor{good}{13.6} & \textcolor{good}{2.30x} & 76.60 & 77.66 & 82.47\\
\hline
\hline
\multirow{9}{*}{\textbf{MobileNet-V2}} & \textbf{BASE} & \textbf{9.4} & \textbf{1.00x} &  \textbf{87.80} & - & - \\
& CG=16, PR=0.05 & \textcolor{bad}{8.0} & \textcolor{bad}{1.18x} & 78.10 &  81.70 & 87.02 \\
& CG=32, PR=0.05 & 7.1 & 1.32x & \textcolor{good}{79.38} & \textcolor{good}{82.62} & \textcolor{good}{87.95} \\
& CG=16, PR=0.10 & 7.4 & 1.27x & 77.60 & 80.70 & 86.48 \\
& \textbf{\textit{CG=32, PR=0.10}} & \textbf{\textit{6.8}} & \textbf{\textit{1.38x}} & \textbf{\textit{78.82}} & \textbf{\textit{81.76}} & \textbf{\textit{87.41}} \\
& CG=16, PR=0.20 & 5.9 & 1.59x & 74.04 & 76.66 & 82.14 \\
& CG=32, PR=0.20 & 5.8 & 1.62x & 77.32 & 80.10 & 85.44 \\
& CG=16, PR=0.30 & 4.6 & 2.04x & \textcolor{bad}{67.82} & \textcolor{bad}{69.86} & \textcolor{bad}{75.86} \\
& CG=32, PR=0.30 & \textcolor{good}{4.4} & \textcolor{good}{2.14x} & 72.36 & 74.64 & 80.77 \\
\hline
\end{tabular}
\end{table*}

For ResNet-152, the configuration \gls{cg}=32 and \gls{pr}=0.30 was selected for discussion in the main text. Despite being the most aggressive pruning configuration evaluated for this architecture, it achieves the highest compression rate (2.37x) while incurring only a modest 1.86\% accuracy drop relative to the baseline. This highlights the substantial over-parameterization of larger models and their ability to withstand aggressive pruning with limited performance degradation.

For EfficientNet-B2, we selected the configuration \gls{cg}=32 and \gls{pr}=0.05, as it maintains higher predictive accuracy while still providing meaningful model compression. For MobileNet-V2, the configuration \gls{cg}=32 and \gls{pr}=0.10 was chosen following the same trade-off criterion, avoiding both the conservative \gls{pr}=0.05 configurations and the excessive accuracy loss observed at \gls{pr}=0.30.

Bold type in Table~\ref{tab:pruning_configs} indicates the selected pruning configurations for each architecture. Overall, these results indicate that the effectiveness of structured pruning is strongly architecture-dependent. That careful selection of pruning parameters enables significant model size reduction with minimal impact on performance, particularly for larger, over-parameterized architectures such as ResNet-152.

Values highlighted in green indicate the best results for each metric within each architecture, while those highlighted in red correspond to the worst-performing configurations. Notably, these worst-case configurations consistently correspond to the highest pruning ratios applied to compact architectures such as MobileNet-V2 and EfficientNet-B2, where aggressive pruning leads to substantial accuracy degradation due to their limited parameter redundancy. In contrast, larger architectures like ResNet-152 exhibit greater resilience, tolerating higher pruning ratios before suffering comparable performance losses.

When selecting the pruning configuration for the main experiments, a trade-off between these two extremes was deliberately considered, with practical deployments on edge devices in mind. Rather than choosing the most aggressive configuration in terms of compression or the one that maximizes accuracy alone, the selected models aim to balance model size reduction and predictive performance.

\end{document}